\documentclass[10pt,twocolumn,letterpaper]{article}

\usepackage{cvpr}
\usepackage{times}
\usepackage{epsfig}
\usepackage{graphicx}
\usepackage{amsmath}
\usepackage{amssymb}
\usepackage{subcaption}
\usepackage{multirow}
\usepackage{color}

\usepackage[pagebackref=true,breaklinks=true,letterpaper=true,colorlinks,bookmarks=false]{hyperref}
\usepackage{cleveref}

\cvprfinalcopy 

\def\cvprPaperID{7280} 

\begin{document}

\title{Action Segmentation with Joint Self-Supervised Temporal Domain Adaptation}

\author{Min-Hung Chen$^1$\thanks{Work done during an internship at Baidu USA} \hspace{1em}
Baopu Li$^2$ \quad
Yingze Bao$^2$ \quad
Ghassan AlRegib$^1$ \quad
Zsolt Kira$^1$\\
$^1$Georgia Institute of Technology \quad
$^2$Baidu USA\\
}

\maketitle

\begin{abstract}
Despite the recent progress of fully-supervised action segmentation techniques, the performance is still not fully satisfactory. One main challenge is the problem of spatio-temporal variations (e.g. different people may perform the same activity in various ways).
Therefore, we exploit unlabeled videos to address this problem by reformulating the action segmentation task as a cross-domain problem with domain discrepancy caused by spatio-temporal variations. To reduce the discrepancy, we propose \textbf{Self-Supervised Temporal Domain Adaptation (SSTDA)}, which contains two self-supervised auxiliary tasks (binary and sequential domain prediction) to jointly align cross-domain feature spaces embedded with local and global temporal dynamics, achieving better performance than other Domain Adaptation (DA) approaches.
On three challenging benchmark datasets (GTEA, 50Salads, and Breakfast), SSTDA outperforms the current state-of-the-art method by large margins (e.g. for the F1$@25$ score, from 59.6\% to 69.1\% on Breakfast, from 73.4\% to 81.5\% on 50Salads, and from 83.6\% to 89.1\% on GTEA), and requires only 65\% of the labeled training data for comparable performance, demonstrating the usefulness of adapting to unlabeled target videos across variations. 
The source code is available at \url{https://github.com/cmhungsteve/SSTDA}.
\end{abstract}

\section{Introduction}
\begin{figure}[!t]
\centering
\includegraphics[width=0.475\textwidth]{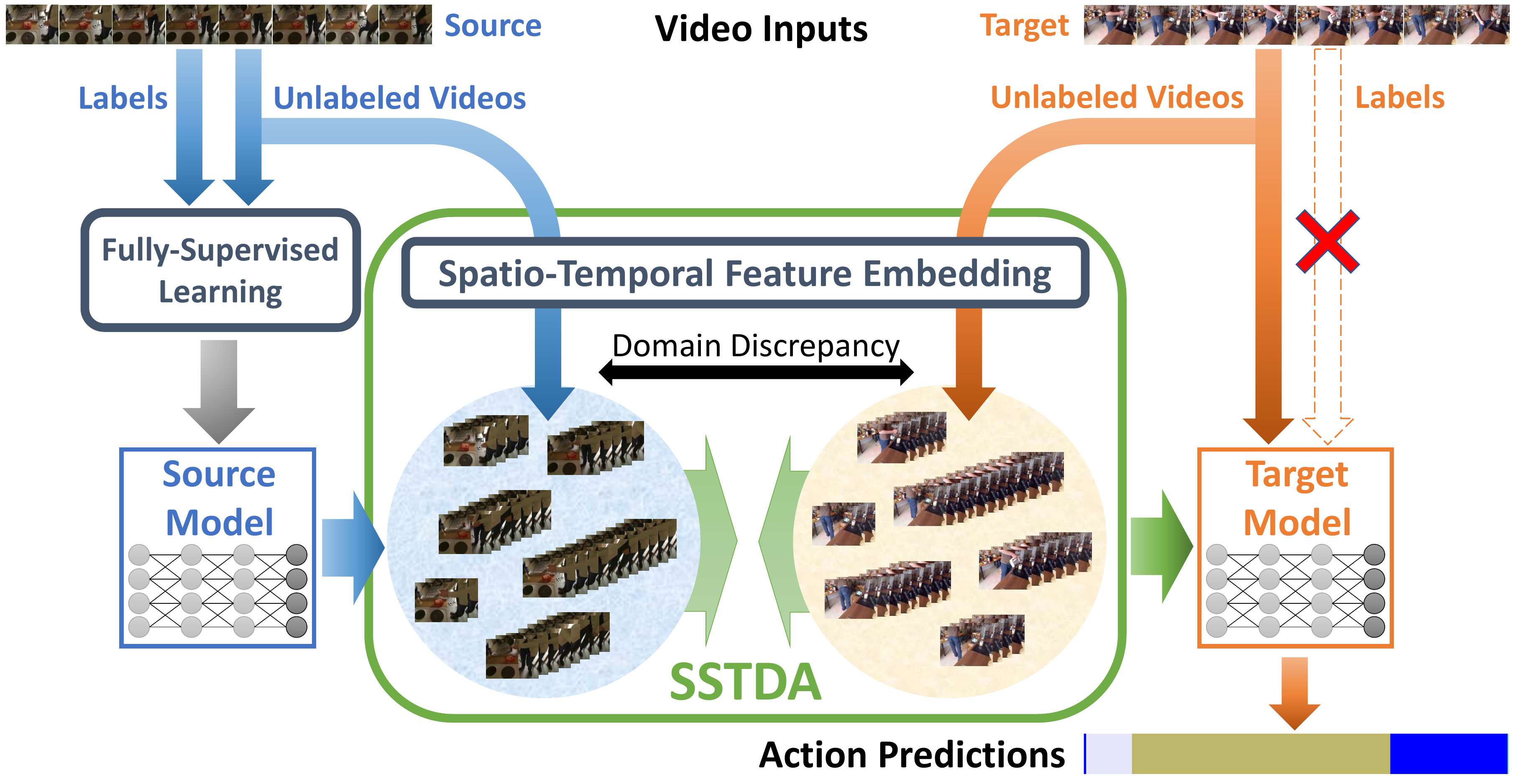}
\caption{An overview of the proposed Self-Supervised Temporal Domain Adaptation (SSTDA) for action segmentation. ``Source" refers to the data with labels, and ``Target" refers to the data without access to labels. SSTDA can effectively adapts the source model trained with standard fully-supervised learning to a target domain by 
diminishing the discrepancy of embedded feature spaces between the two domains caused by spatio-temporal variations. SSTDA only requires unlabeled videos from both domains with the standard transductive setting, which eliminates the need of additional labels to obtain the final target model.
}
\label{fig:overview_action_seg_DA}
\end{figure}

The goal of action segmentation is to simultaneously segment videos by time and predict an action class for each segment, leading to various applications (e.g. human activity analyses).
While action classification has shown great progress given the recent success of deep neural networks~\cite{wang2018non, ma2018attend, ma2019ts}, temporally locating and recognizing action segments in long videos is still challenging.
One main challenge is the problem of \textit{spatio-temporal variations} of human actions across videos~\cite{kong2018human}. For example, different people may \textit{make tea} in different personalized styles even if the given recipe is the same. The intra-class variations cause degraded performance by directly deploying a model trained with different groups of people. 

Despite significant progress made by recent methods based on temporal convolution with fully-supervised learning~\cite{lea2017temporal, ding2017tricornet, lei2018temporal, farha2019ms}, the performance is still not fully satisfactory (e.g. the best accuracy on the Breakfast dataset is still lower than 70\%). One method to improve the performance is to exploit knowledge from larger-scale labeled data~\cite{carreira2017quo}.
However, manually annotating precise frame-by-frame actions is time-consuming and challenging. Another way is to design more complicated architectures but with higher costs of model complexity.
Thus, we aim to address the spatio-temporal variation problem with unlabeled data, which are comparatively easy to obtain. To achieve this goal, we propose to diminish the distributional discrepancy caused by spatio-temporal variations by exploiting auxiliary unlabeled videos with the same types of human activities performed by different people. 
More specifically, to extend the framework of the main video task for exploiting auxiliary data~\cite{zhang2019learning,lahiri2019unsupervised}, we reformulate our main task as an unsupervised domain adaptation (DA) problem with the transductive setting~\cite{pan2010survey, csurka2017comprehensive}, which aims to reduce the discrepancy between source and target domains without access to the target labels.

Recently, adversarial-based DA approaches~\cite{ganin2015unsupervised, ganin2016domain,tzeng2017adversarial,zhang2018collaborative} show progress in reducing the discrepancy for images using a domain discriminator equipped with adversarial training. However, videos also suffer from domain discrepancy along the temporal direction~\cite{chen2019temporal}, so using image-based domain discriminators is not sufficient for action segmentation. 
Therefore, we propose \textbf{Self-Supervised Temporal Domain Adaptation (SSTDA)}, containing two self-supervised auxiliary tasks: 1) \textit{binary domain prediction}, which predicts a single domain for each frame-level feature, and 2) \textit{sequential domain prediction}, which predicts the permutation of domains for an untrimmed video. Through adversarial training with both auxiliary tasks, SSTDA can jointly align cross-domain feature spaces that embed local and global temporal dynamics, to address the spatio-temporal variation problem for action segmentation, as shown in \Cref{fig:overview_action_seg_DA}. 
To support our claims, we compare our method with other popular DA approaches and show better performance, demonstrating the effectiveness for aligning temporal dynamics by SSTDA.
Finally, we evaluate our approaches on three datasets with high spatio-temporal variations: GTEA~\cite{fathi2011learning}, 50Salads~\cite{stein2013combining}, and the Breakfast dataset~\cite{kuehne2014language}. By exploiting unlabeled target videos with SSTDA, our approach outperforms the current state-of-the-art methods by large margins and achieve comparable performance using only \textit{65\%} of labeled training data. 

In summary, our contributions are three-fold:
\begin{enumerate}
    \item \textbf{Self-Supervised Sequential Domain Prediction}: 
    We propose a novel self-supervised auxiliary task, which predicts the permutation of domains for long videos, to facilitate video domain adaptation. To the best of our knowledge, this is the first self-supervised method designed for cross-domain action segmentation.
    
    \item \textbf{Self-Supervised Temporal Domain Adaptation (SSTDA)}:
    By integrating two self-supervised auxiliary tasks, \textit{binary} and \textit{sequential domain prediction}, our proposed SSTDA can jointly align local and global embedded feature spaces across domains, outperforming other DA methods. 
    
    \item \textbf{Action Segmentation with SSTDA}:
    By integrating SSTDA for action segmentation, our approach outperforms the current state-of-the-art approach by large margins, and achieve comparable performance by using only \textit{65\%} of labeled training data. Moreover, different design choices are analyzed to identify the key contributions of each component.
    
\end{enumerate}

\section{Related Works} 

\noindent\textbf{Action Segmentation}
methods proposed recently are built upon temporal convolution networks (TCN)~\cite{lea2017temporal, ding2017tricornet, lei2018temporal, farha2019ms} because of their ability to capture long-range dependencies across frames and faster training compared to RNN-based methods. 
With the multi-stage pipeline, MS-TCN~\cite{farha2019ms} performs hierarchical temporal convolutions to effectively extract temporal features and achieve the state-of-the-art performance for action segmentation. In this work, we utilize MS-TCN as the baseline model and integrate the proposed self-supervised modules to further boost the performance \emph{without extra labeled data}.

\noindent\textbf{Domain Adaptation (DA)}
has been popular recently especially with the integration of deep learning. With the two-branch (source and target) framework for most DA works, finding a common feature space between source and target domains is the ultimate goal, and the key is to design the domain loss to achieve this goal~\cite{csurka2017comprehensive}.

\textit{Discrepancy-based DA}~\cite{long2015learning,long2016unsupervised,long2017deep} is one of the major classes of methods where the main goal is to reduce the distribution distance between the two domains.
\textit{Adversarial-based DA}~\cite{ganin2015unsupervised, ganin2016domain} is also popular with similar concepts as GANs~\cite{goodfellow2014generative} by using domain discriminators. With carefully designed adversarial objectives, the domain discriminator and the feature extractor are optimized through min-max training.
Some works further improve the performance by assigning pseudo-labels to target data~\cite{pei2018multi,xie2018learning}.
Furthermore, \textit{Ensemble-based DA}~\cite{saito2018maximum,lee2019sliced} incorporates multiple target branches to build an ensemble model. 
Recently, \textit{Attention-based DA}~\cite{wang2019transferable,kurmi2019attending} assigns attention weights to different regions of images for more effective DA. 

Unlike images, video-based DA is still under-explored. Most works concentrate on small-scale video DA datasets~\cite{sultani2014human, xu2016dual, jamal2018deep}. 
Recently, two larger-scale cross-domain video classification datasets along with the state-of-the-art approach are proposed~\cite{chen2019taaan,chen2019temporal}. 
Moreover, some authors also proposed novel frameworks to utilize auxiliary data for other video tasks, including object detection~\cite{lahiri2019unsupervised} and action localization~\cite{zhang2019learning}. 
These works differ from our work by either different video tasks~\cite{lahiri2019unsupervised,chen2019taaan,chen2019temporal} or access to the labels of auxiliary data~\cite{zhang2019learning}.

\noindent\textbf{Self-Supervised Learning}
has become popular in recent years for images and videos given the ability to learn informative feature representations without human supervision. The key is to design an auxiliary task (or pretext task) that is related to the main task and the labels can be self-annotated. Most of the recent works for videos design auxiliary tasks based on spatio-temporal orders of videos~\cite{lee2017unsupervised,wei2018learning,kim2019self,ahsan2019video,xu2019self}. Different from these works, our proposed auxiliary task predicts temporal permutation for cross-domain videos, aiming to address the problem of spatio-temporal variations for action segmentation.

\section{Technical Approach}
In this section, the baseline model which is the current state-of-the-art for action segmentation, MS-TCN~\cite{farha2019ms}, is reviewed first (\Cref{sec:baseline}). Then the novel temporal domain adaptation scheme consisting of two self-supervised auxiliary tasks, binary domain prediction (\Cref{sec:LTDA}) and sequential domain prediction (\Cref{sec:GTDA}), is proposed, followed by the final action segmentation model.

\begin{figure}[!t]
\centering
\includegraphics[width=0.475\textwidth]{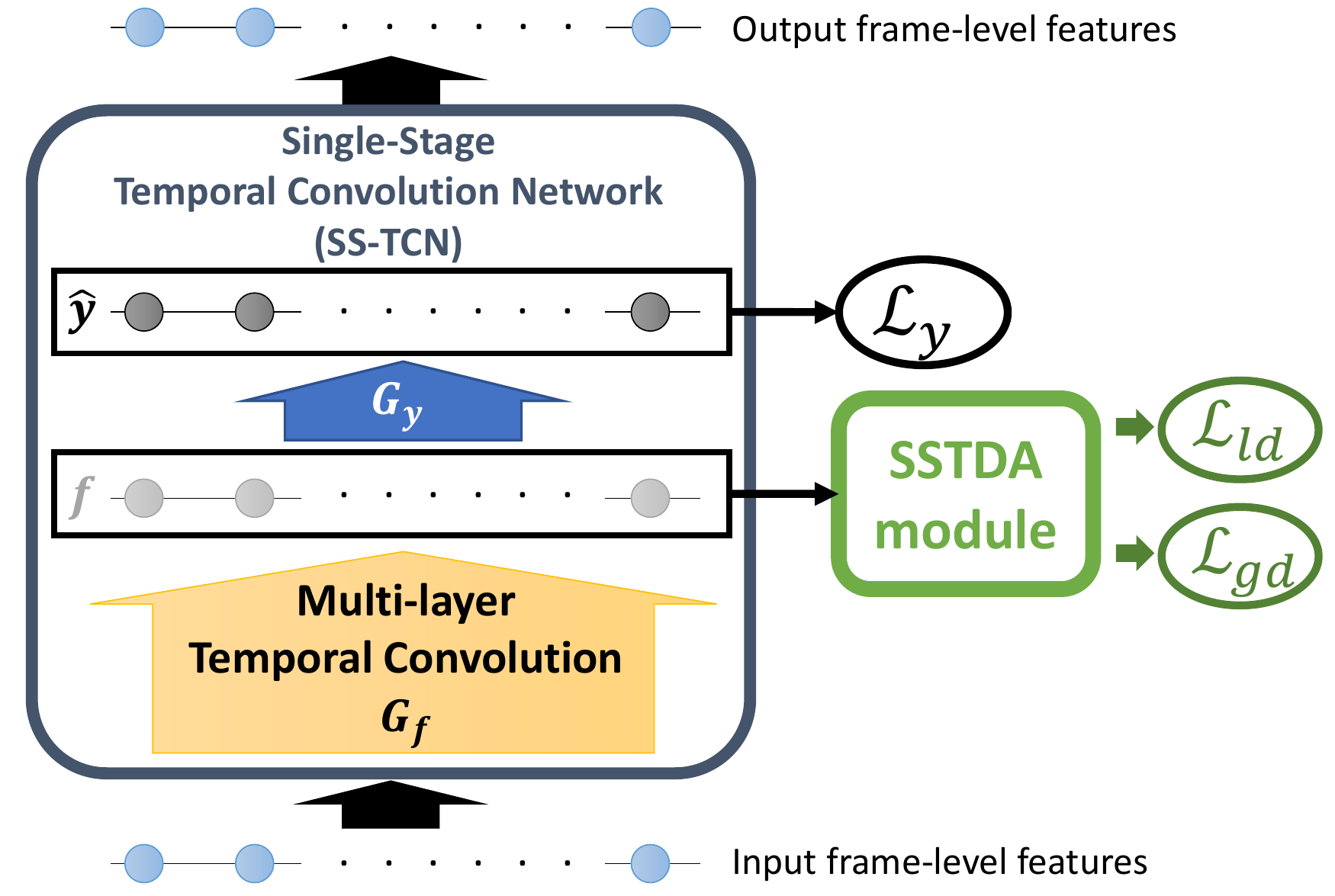}
\caption{Illustration of the baseline model and the integration with our proposed SSTDA. The frame-level features $\textit{\textbf{f}}$ are obtained by applying 
the temporal convolution network 
${G}_{f}$ to the inputs, and converted to the corresponding predictions $\mathbf{\hat{y}}$ using a fully-connected layer $G_y$ to calculate the prediction loss $\mathcal{L}_{y}$. The SSTDA module is integrated with $\textit{\textbf{f}}$ to calculate the local and global domain losses, $\mathcal{L}_{ld}$ and $\mathcal{L}_{gd}$ for optimizing $\textit{\textbf{f}}$ during training (see details in \Cref{sec:SSTDA}).
Here we only show one stage in our multi-stage model.}
\label{fig:full_arch_single_stage}
\end{figure}

\subsection{Baseline Model} \label{sec:baseline}
Our work is built on the current state-of-the-art model for action segmentation, multi-stage temporal convolutional network (MS-TCN)~\cite{farha2019ms}. For each stage, a single-stage TCN (SS-TCN) applies a multi-layer TCN, $G_f$, to derive the frame-level features $\textit{\textbf{f}}=\{f_1, f_2, ..., f_T\}$, and makes the corresponding predictions $\mathbf{\hat{y}}=\{{\hat{y}}_1, {\hat{y}}_2, ..., {\hat{y}}_T\}$ using a fully-connected layer $G_y$. By following \cite{farha2019ms}, the prediction loss $\mathcal{L}_{y}$ is calculated based on the predictions $\mathbf{\hat{y}}$, as shown in the left part of \Cref{fig:full_arch_single_stage}. 
Finally, multiple stages of SS-TCNs are stacked to enhance the temporal receptive fields, constructing the final baseline model, MS-TCN, where each stage takes the predictions from the previous stage as inputs, and makes predictions for the next stage.

\subsection{Self-Supervised Temporal Domain Adaptation} \label{sec:SSTDA}
Despite the promising performance of MS-TCN on action segmentation over previous methods, there is still a large room for improvement. One main challenge is the problem of \textit{spatio-temporal variations} of human actions~\cite{kong2018human}, causing the distributional discrepancy across domains~\cite{csurka2017comprehensive}. 
For example, different subjects may perform the same action completely differently due to personalized spatio-temporal styles. 
Moreover, collecting annotated data for action segmentation is challenging and time-consuming. Thus, such challenges motivate the need to learn domain-invariant feature representations without full supervision. 
Inspired by the recent progress of self-supervised learning, which learns informative features that can be transferred to the main target tasks without external supervision (e.g. human annotation), we propose \textbf{Self-Supervised Temporal Domain Adaptation (SSTDA)} to diminish cross-domain discrepancy by designing self-supervised auxiliary tasks using unlabeled videos.

To effectively transfer knowledge, the self-supervised auxiliary tasks should be closely related to the main task, which is cross-domain action segmentation in this paper. 
Recently, adversarial-based DA approaches~\cite{ganin2015unsupervised, ganin2016domain} show progress in addressing cross-domain image problems using a domain discriminator with adversarial training where domain discrimination can be regarded as a self-supervised auxiliary task since domain labels are self-annotated.  
However, directly applying image-based DA for video tasks results in sub-optimal performance due to the temporal information being ignored~\cite{chen2019temporal}.
Therefore, the question becomes: \textit{How should we design the self-supervised auxiliary tasks to benefit cross-domain action segmentation?} More specifically, the answer should address both \textit{cross-domain} and \textit{action segmentation} problems.

To address this question, we first apply an auxiliary task \textit{binary domain prediction} to predict the domain for each frame where the frame-level features are embedded with local temporal dynamics, aiming to address the cross-domain problems for videos in local scales. Then we propose a novel auxiliary task \textit{sequential domain prediction} to temporally segment domains for untrimmed videos where the video-level features are embedded with global temporal dynamics, aiming to fully address the above question. Finally, SSTDA is achieved locally and globally by jointly applying these two auxiliary tasks, as illustrated in \Cref{fig:SSTDA_intuition}.

In practice, since the key for effective video DA is to simultaneously align and learn temporal dynamics, instead of separating the two processes~\cite{chen2019temporal}, we integrate SSTDA modules to multiple stages instead of the last stage only, and the single-stage integration is illustrated in \Cref{fig:full_arch_single_stage}. 

\subsubsection{Local SSTDA} \label{sec:LTDA}
\begin{figure}[!t]
\centering
\includegraphics[width=0.475\textwidth]{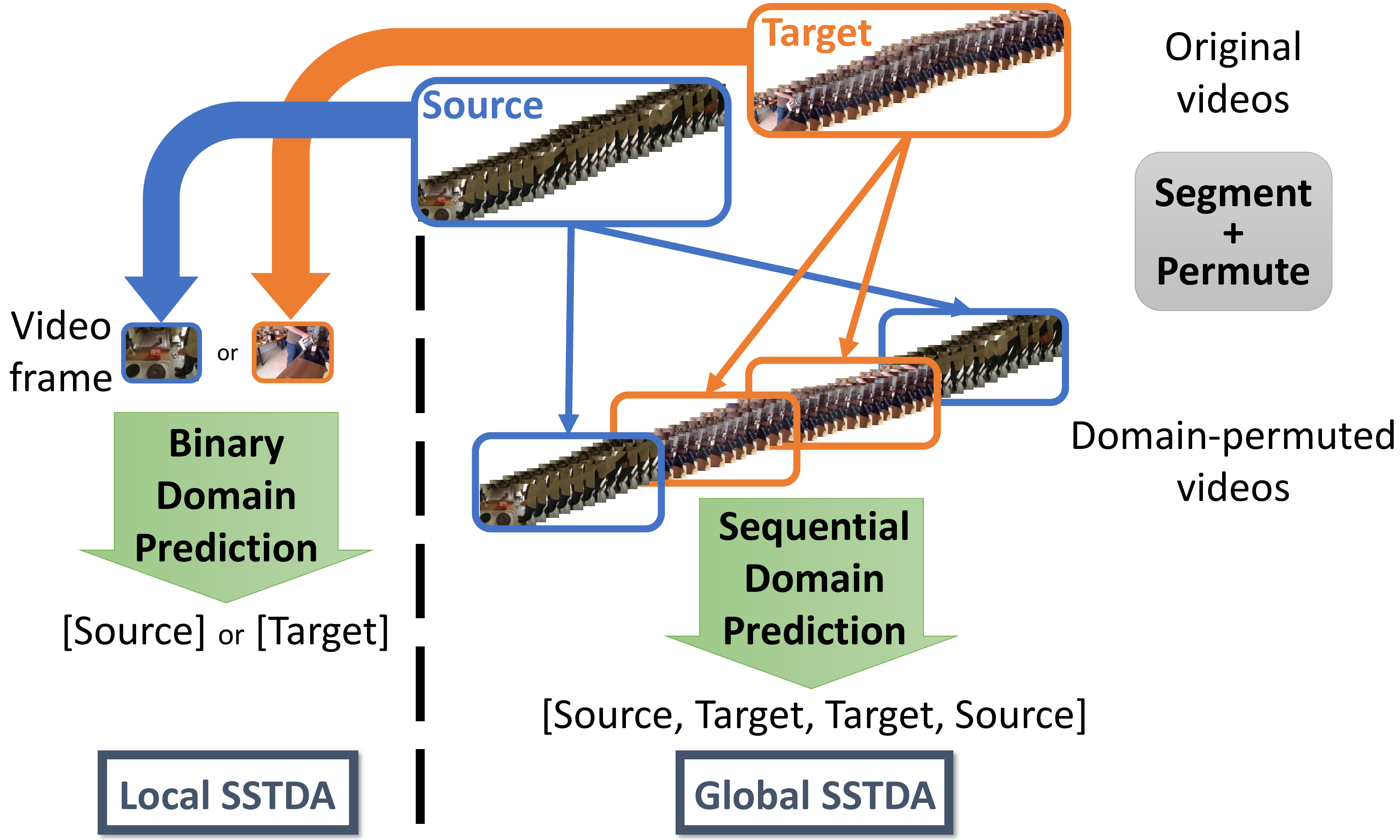}
\caption{The two self-supervised auxiliary tasks in SSTDA: 1) \textit{binary domain prediction}: discriminate single frame, 2) \textit{sequential domain prediction}: predict a sequence of domains for an untrimmed video. These two tasks contribute to local and global SSTDA, respectively.}
\label{fig:SSTDA_intuition}
\end{figure}
The main goal of action segmentation is to learn frame-level feature representations that encode spatio-temporal information so that the model can exploit information from multiple frames to predict the action for each frame. 
Therefore, we first learn domain-invariant frame-level features with the auxiliary task \textit{binary domain prediction} (\Cref{fig:SSTDA_intuition} left).

\noindent\textbf{Binary Domain Prediction:} 
For a single stage, we feed the frame-level features from source and target domains $\boldsymbol{f^S}$ and $\boldsymbol{f^T}$, respectively, to an additional shallow \textit{binary domain classifier} $G_{ld}$, to discriminate which domain the features come from. Since temporal convolution from previous layers encodes information from multiple adjacent frames to each frame-level feature, those frames contribute to the binary domain prediction for each frame.
Through adversarial training with a gradient reversal layer (GRL)~\cite{ganin2015unsupervised, ganin2016domain}, which reverses the gradient signs during back-propagation, $G_f$ will be optimized to gradually align the feature distributions between the two domains. Here we note $\hat{G}_{ld}$ as $G_{ld}$ equipped with GRL, as shown in \Cref{fig:SSTDA_full}. 

Since this work is built on MS-TCN, \textit{integrating $\hat{G}_{ld}$ with proper stages} is critical for effective DA.
From our investigation, the best performance happens when $\hat{G}_{ld}$s are integrated into middle stages. See \Cref{sec:design_local_SSTDA} for details.

The overall loss function becomes a combination of the baseline prediction loss $\mathcal{L}_y$ and the local domain loss $\mathcal{L}_{ld}$ with reverse sign, which can be expressed as follows:
\begin{equation} \label{eq:loss_baseline-localdomain}
\small
\begin{split}
\mathcal{L} = \sum^{N_s}\mathcal{L}_y - \sum^{\widetilde{N_s}}\beta_l\mathcal{L}_{ld}
\end{split}
\end{equation}
\begin{equation} \label{eq:loss_localdomain}
\small
\mathcal{L}_{ld} = \frac{1}{T}\sum_{j=1}^{T}L_{ld}(G_{ld}(f_j),d_j)
\end{equation}
where $N_s$ is the total stage number in MS-TCN, $\widetilde{N_s}$ is the number of stages integrated with $\hat{G}_{ld}$, and $T$ is the total frame number of a video. $\textit{L}_{ld}$ is a binary cross-entropy loss function, and $\beta_l$ is the trade-off weight for local domain loss $\mathcal{L}_{ld}$, obtained by following the common strategy as \cite{ganin2015unsupervised,ganin2016domain}.

\subsubsection{Global SSTDA} \label{sec:GTDA}
\begin{figure*}[!t]
\centering
\includegraphics[width=\textwidth]{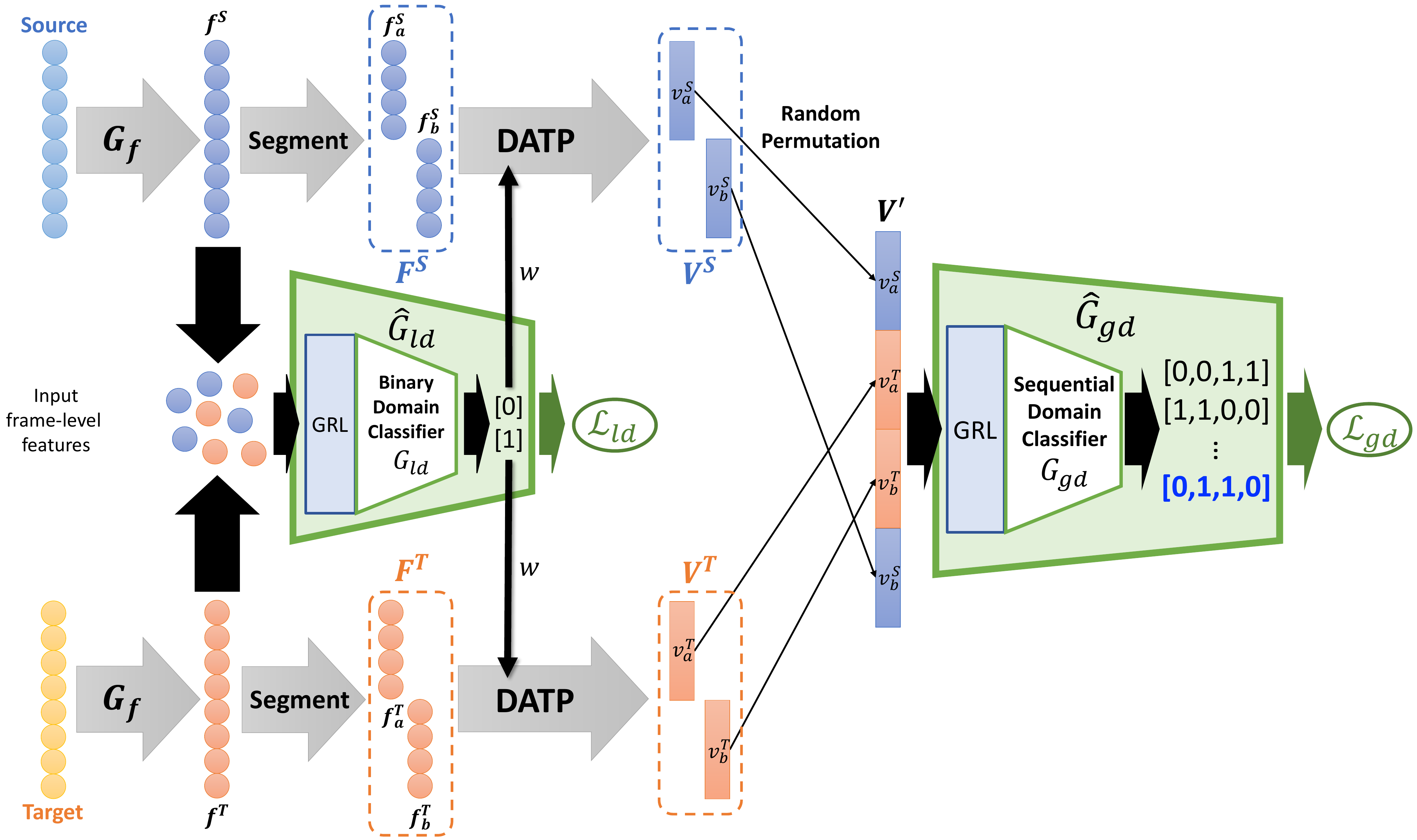}
\caption{The overview of the proposed Self-Supervised Temporal Domain Adaptation (SSTDA). The inputs from the two domains are first encoded with local temporal dynamics using $G_{f}$ to obtain the frame-level features $\boldsymbol{f^S}$ and $\boldsymbol{f^T}$, respectively. We apply local SSTDA on all $\boldsymbol{f}$ using binary domain prediction $\hat{G}_{ld}$. Besides, $\boldsymbol{f^S}$ and $\boldsymbol{f^T}$ are evenly divided into multiple segments to learn segment-level features $\boldsymbol{V^S}$ and $\boldsymbol{V^T}$ by DATP, respectively. Finally, the global SSTDA is applied on $\boldsymbol{V'}$, which is generated by concatenating shuffled $\boldsymbol{V^S}$ and $\boldsymbol{V^T}$, using sequential domain prediction $\hat{G}_{gd}$. $\mathcal{L}_{ld}$ and $\mathcal{L}_{gd}$
are the domain losses from $\hat{G}_{ld}$ and $\hat{G}_{gd}$, respectively. $w$ corresponds to the attention weights for DATP, which are calculated form the outputs of $\hat{G}_{ld}$. Here we use 8-frame videos and 2 segments as an example for this figure. Best views in colors.}
\label{fig:SSTDA_full}
\end{figure*}

Although frame-level features $\textit{\textbf{f}}$ is learned using the context and dependencies from neighbor frames, the temporal receptive fields of $\textit{\textbf{f}}$ are still limited, unable to represent full videos. Solely integrating DA into $\textit{\textbf{f}}$ cannot fully address spatio-temporal variations for untrimmed long videos. Therefore, in addition to binary domain prediction for frame-level features, we propose the second self-supervised auxiliary task for video-level features: \textbf{sequential domain prediction}, which predicts a sequence of domains for video clips, as shown in the right part of \Cref{fig:SSTDA_intuition}. This task is a temporal domain segmentation problem, aiming to predict the correct permutation of domains for long videos consisting of shuffled video clips from both source and target domains. 
Since this goal is related to both cross-domain and action segmentation problems, \textit{sequential domain prediction} can effectively benefit our main task.

More specifically, we first divide $\boldsymbol{f^S}$ and $\boldsymbol{f^T}$ into two sets of segments $\boldsymbol{F^S}=\{\boldsymbol{f^S_a},\boldsymbol{f^S_b},...\}$ and $\boldsymbol{F^T}=\{\boldsymbol{f^T_a},\boldsymbol{f^T_b},...\}$, respectively, and then learn the corresponding two sets of segment-level feature representations $\boldsymbol{V^S}=\{v^S_a,v^S_b,...\}$ and $\boldsymbol{V^T}=\{v^T_a,v^T_b,...\}$ with \textit{Domain Attentive Temporal Pooling (DATP)}. All features $v$ are then shuffled and combined in random order and fed to a \textit{sequential domain classifier $G_{gd}$} equipped with GRL (noted as $\hat{G}_{gd}$) to predict the permutation of domains, as shown in \Cref{fig:SSTDA_full}.

\noindent\textbf{Domain Attentive Temporal Pooling (DATP):}
The most straightforward method to obtain a video-level feature is to aggregate frame-level features using \textit{temporal pooling}. However, not all the frame-level features contribute the same to the overall domain discrepancy, as mentioned in \cite{chen2019temporal}. Hence, we assign larger attention weights $w_j$ (calculated using $\hat{G}_{gd}$ in local SSTDA) to the features which have larger domain discrepancy so that we can focus more on aligning those features.
Finally, the attended frame-level features are aggregated with temporal pooling to generate the video-level feature $v$, which can be expressed as:
\begin{equation} \label{eq:attention-residual}
\small
v = \frac{1}{T'}\sum_{j=1}^{T'}w_j \cdot f_j
\end{equation}
where $T'$ is the number of frames in a video segment. For more details, please refer to the supplementary.

\noindent\textbf{Sequential Domain Prediction:} 
By separately applying DATP to both source and target segments, respectively, a set of segment-level feature representations $\boldsymbol{V}=\{v^S_a,v^S_b,...,v^T_a,v^T_b,...\}$ are obtained.
We then shuffle all the features in $\boldsymbol{V}$ and concatenate them into a feature to represent a long and untrimmed video $\boldsymbol{V'}$, which contains video segments from both domains in random order. Finally, $\boldsymbol{V'}$ is fed into a \textit{sequential domain classifier $G_{gd}$} to predict the permutation of domains for the video segments. For example, if $\boldsymbol{V'}=[v^S_a,v^T_a,v^T_b,v^S_b]$, the goal of $G_{gd}$ is to predict the permutation as $[0,1,1,0]$. $G_{gd}$ is a multi-class classifier where the class number corresponds to the total number of all possible permutations of domains, and the complexity of $G_{gd}$ is determined by the segment number for each video (more analyses in \Cref{sec:design_global_SSTDA}). 
The outputs of $G_{gd}$ are used to calculate the global domain loss $\mathcal{L}_{gd}$ as below:
\begin{equation} \label{eq:loss_globaldomain}
\small
\mathcal{L}_{gd} = L_{gd}(G_{gd}(\boldsymbol{V'})),y_d)
\end{equation}
where $\textit{L}_{gd}$ is also a standard cross-entropy loss function where the class number is determined by the segment number.
Through adversarial training with GRL, \textit{sequential domain prediction} also contributes to optimizing $G_f$ to align the feature distributions between the two domains. 

There are some self-supervised learning works also proposing the concepts of \textit{temporal shuffling}~\cite{lee2017unsupervised,xu2019self}. However, they predict temporal orders within one domain, aiming to learn general temporal information for video features. Instead, our method predicts temporal permutation for cross-domain videos, which are shown with a dual-branch pipeline in \Cref{fig:SSTDA_full}, and integrate with binary domain prediction to effectively address both \textit{cross-domain} and \textit{action segmentation} problems.

\subsubsection{Local-Global Joint Training.} \label{sec:local_global}
Finally, we also adopt a strategy from \cite{wang2019transferable} to minimize the class entropy for the frames that are similar across domains by adding a domain attentive entropy (DAE) loss $\mathcal{L}_{ae}$. Please refer to the supplementary for more details.

By adding the global domain loss $\mathcal{L}_{gd}$ (\Cref{eq:loss_globaldomain}) and the attentive entropy loss $\mathcal{L}_{ae}$ into \Cref{eq:loss_baseline-localdomain}, the overall loss of our final proposed \textbf{Self-Supervised Temporal Domain Adaptation (SSTDA)} can be expressed as follows:
\begin{equation} \label{eq:loss_SSTDA}
\small
\begin{split}
\mathcal{L} = \sum^{N_s}\mathcal{L}_y - \sum^{\widetilde{N_s}}(\beta_l\mathcal{L}_{ld}+\beta_g\mathcal{L}_{gd}-\mu\mathcal{L}_{ae})
\end{split}
\end{equation}
where $\beta_g$ and $\mu$ are the weights for $\mathcal{L}_{gd}$ and $\mathcal{L}_{ae}$, respectively.

\section{Experiments}
To validate the effectiveness of the proposed methods in reducing spatial-temporal discrepancy for action segmentation, we choose three challenging datasets: 
GTEA~\cite{fathi2011learning}, 50Salads~\cite{stein2013combining}, and Breakfast~\cite{kuehne2014language}, which separate the training and validation sets by different people (noted as \textit{subjects}) with leave-subjects-out cross-validation for evaluation, resulting in large domain shift problem due to spatio-temporal variations. 
Therefore, we regard the training set as \textit{Source} domain, and the validation set as \textit{Target} domain with the standard transductive unsupervised DA protocol~\cite{pan2010survey, csurka2017comprehensive}.
See the supplementary for more implementation details.

\subsection{Datasets and Evaluation Metrics}
\begin{table}[!t]
\centering
    \begin{tabular}{c|c|c|c}
     & \textbf{GTEA} & \textbf{50Salads} & \textbf{Breakfast} \\ \hline
    subject \# & 4 & 25 & 52 \\ \hline
    class \# & 11 & 17 & 48 \\ \hline
    video \# & 28 & 50 & 1712 \\ \hline
    leave-\#-subject-out & 1 & 5 & 13 \\ \hline
    \end{tabular}
\caption{The statistics of action segmentation datasets.}
\label{table:dataset}
\end{table}
The overall statistics of the three datasets are listed in \Cref{table:dataset}. 
Three widely used evaluation metrics are chosen as follows \cite{lea2017temporal}: frame-wise \textit{accuracy (Acc)}, segmental \textit{edit score}, and segmental F1 score at the IoU threshold $k\%$, denoted as \textit{F1$@k$ ($k=\{10,25,50\}$)}. While \textit{Acc} is the most common metric, \textit{edit} and \textit{F1 score} both consider the temporal relation between predictions and ground truths, better reflecting the performance for action segmentation.

\subsection{Experimental Results}
We first investigate the effectiveness of our approaches in utilizing unlabeled target videos for action segmentation. 
We choose MS-TCN~\cite{farha2019ms} as the backbone model since it is the current state of the art for this task.
``Source only" means the model is trained only with source labeled videos, i.e., the baseline model. And then our approach is compared to other methods with the same transductive protocol. Finally, we compare our method to the most recent action segmentation methods on all three datasets, and investigate how our method can reduce the reliance on source labeled data.

\noindent\textbf{Self-Supervised Temporal Domain Adaptation:}
First we investigate the performance of local SSTDA by integrating the auxiliary task \textit{binary domain prediction} with the baseline model.
The results on all three datasets are improved significantly, as shown in \Cref{table:exp_local_global_DA}. For example, on the GTEA dataset, our approach outperforms the baseline by 4.3\% for F1$@$25, 3.2\% for the edit score and 3.6\% for the frame-wise accuracy. Although local SSTDA mainly works on the frame-level features, the temporal information is still encoded using the context from neighbor frames, helping address the variation problem for videos across domains.

\begin{table}[!t]
\centering
\small
    \begin{tabular}{c|ccc|c|c} 
    \hline
    \textbf{GTEA} & \multicolumn{3}{c|}{F1$@\{10,25,50\}$} & Edit & Acc \\ \hline
    Source only (MS-TCN)$\dag$ & 86.5 & 83.6 & 71.9 & 81.3 & 76.5 \\ 
    Local SSTDA & 89.6 & 87.9 & 74.4 & 84.5 & \textbf{80.1} \\ 
    SSTDA$\ddag$ & \textbf{90.0} & \textbf{89.1} & \textbf{78.0} & \textbf{86.2} & 79.8 \\ 
    \hline
    \hline
    \textbf{50Salads} & \multicolumn{3}{c|}{F1$@\{10,25,50\}$} & Edit & Acc \\ \hline
    Source only (MS-TCN)$\dag$ & 75.4 & 73.4 & 65.2 & 68.9 & 82.1 \\ 
    Local SSTDA & 79.2 & 77.8 & 70.3 & 72.0 & 82.8 \\ 
    SSTDA$\ddag$ & \textbf{83.0} & \textbf{81.5} & \textbf{73.8} & \textbf{75.8} & \textbf{83.2} \\ 
    \hline
    \hline
    \textbf{Breakfast} & \multicolumn{3}{c|}{F1$@\{10,25,50\}$} & Edit & Acc \\ \hline
    Source only (MS-TCN)$\dag$ & 65.3 & 59.6 & 47.2 & 65.7 & 64.7 \\ 
    Local SSTDA & 72.8 & 67.8 & 55.1 & 71.7 & \textbf{70.3} \\ 
    SSTDA$\ddag$ & \textbf{75.0} & \textbf{69.1} & \textbf{55.2} & \textbf{73.7} & 70.2 \\ 
    \hline
    \end{tabular}
\caption{The experimental results for our approaches on three benchmark datasets.
``SSTDA" refers to the full model while ``Local SSTDA" only contains binary domain prediction.
$\dag$We achieve higher performance than reported in \cite{farha2019ms} when using the released code, so use that as the baseline performance for the whole paper. $\ddag$Global SSTDA requires outputs from local SSTDA, so it is not evaluated alone.} 
\label{table:exp_local_global_DA}
\end{table}

Despite the improvement from local SSTDA, integrating DA into frame-level features cannot fully address the problem of spatio-temporal variations for long videos. Therefore, we integrate our second proposed auxiliary task sequential domain prediction for untrimmed long videos. By jointly training with both auxiliary tasks, SSTDA can jointly align cross-domain feature spaces embedding with local and global temporal dynamics, and further improve over local SSTDA with significant margins. For example, on the 50Salads dataset, it outperforms local SSTDA by 3.8\% for F1$@$10, 3.7\% for F1$@$25, 3.5\% for F1$@$50, and 3.8\% for the edit score, as shown in \Cref{table:exp_local_global_DA}.  

One interesting finding is that local SSTDA contributes to most of the frame-wise accuracy improvement for SSTDA because it focuses on aligning frame-level feature spaces. 
On the other hand, sequential domain prediction benefits aligning video-level feature spaces, contributing to further improvement for the other two metrics,
which consider temporal relation for evaluation.

\noindent\textbf{Learning from Unlabeled Target Videos:}
We also compare SSTDA with other popular approaches~\cite{ganin2016domain,long2017deep,pei2018multi,xie2018learning,saito2018maximum,lee2019sliced,xu2019self} to validate the effectiveness of reducing spatio-temporal discrepancy with the same amount of unlabeled target videos.
For the fair comparison, we integrate all these methods with the same baseline model, MS-TCN. For more implementation details, please refer to the supplementary.

\Cref{table:exp_other_DA} shows that our proposed SSTDA outperforms all the other investigated DA methods in terms of the two metrics that consider temporal relation. We conjecture the main reason is that all these DA approaches are designed for cross-domain image problems. Although they are integrated with frame-level features which encode local temporal dynamics, the limited temporal receptive fields prevent them from fully addressing temporal domain discrepancy. Instead, the \textit{sequential domain prediction} in SSTDA is directly applied to the whole untrimmed video, helping to globally align the cross-domain feature spaces that embed longer temporal dynamics, so that spatio-temporal variations can be reduced more effectively.

We also compare with the most recent video-based self-supervised learning method, \cite{xu2019self}, which can also learn temporal dynamics from unlabeled target videos. However, the performance is even worse than other DA methods, implying that temporal shuffling \textit{within single domain} does not effectively benefit cross-domain action segmentation.

\begin{table}[!t]
\centering
\small
    \begin{tabular}{c|ccc|c} 
    \hline
     & \multicolumn{3}{c|}{F1$@\{10,25,50\}$} & Edit \\ \hline
    Source only (MS-TCN) & 86.5 & 83.6 & 71.9 & 81.3 \\ \hline
    VCOP~\cite{xu2019self} & 87.3 & 85.9 & 70.1 & 82.2 \\ \hline
    DANN~\cite{ganin2016domain} & 89.6 & 87.9 & 74.4 & 84.5 \\ 
    JAN~\cite{long2017deep} & 88.7 & 87.6 & 73.1 & 83.1 \\ 
    MADA~\cite{pei2018multi} & 88.6 & 86.7 & 75.8 & 83.5 \\ 
    MSTN~\cite{xie2018learning} & 89.9 & 88.2 & 75.9 & 84.7 \\ 
    MCD~\cite{saito2018maximum} & 88.1 & 86.3 & 73.4 & 82.7 \\ 
    SWD~\cite{lee2019sliced} & 89.0 & 87.3 & 73.8 & 84.4 \\ \hline
    \textbf{SSTDA} & \textbf{90.0} & \textbf{89.1} & \textbf{78.0} & \textbf{86.2} \\ 
    \hline
    \end{tabular}
\caption{The comparison of different methods that can learn information from unlabeled target videos (on GTEA). All the methods are integrated with the same baseline model MS-TCN for fair comparison. 
Please refer to the supplementary for the results on other datasets.} 
\label{table:exp_other_DA}
\end{table}

\noindent\textbf{Comparison with Action Segmentation Methods:}
Here we compare the recent methods to SSTDA trained with two settings: 1) fully source labels, and 2) weakly source labels.

The first setting means we have labels for all the frames in source videos, and SSTDA outperforms all the previous methods on the three datasets with respect to all evaluation metrics.
For example, SSTDA outperforms currently the state-of-the-art fully-supervised method, MS-TCN~\cite{farha2019ms}, by large margins (e.g. 8.1\% for F1$@$25, 8.6\% for F1$@$50, and 6.9\% for the edit score on 50Salads; 9.5\% for F1$@$25, 8.0\% for F1$@$50, and 8.0\% for the edit score on Breakfast), as demonstrated in \Cref{table:SOTA}. Since no additional labeled data is used, these results indicate how our proposed SSTDA address the spatio-temporal variation problem with unlabeled videos to improve the action segmentation performance.

Given the significant improvement by exploiting unlabeled target videos, it implies the potential to train with fewer number of labeled frames using SSTDA, which is our second setting. In this setting, we drop labeled frames from source domains with uniform sampling for training, and evaluate on the same length of validation data. Our experiment indicates that by integrating with SSTDA, only \textit{65\%} of labeled training data are required to achieve comparable performance with MS-TCN, as shown in the ``SSTDA (65\%)" row in \Cref{table:SOTA}. For the full experiments about labeled data reduction, please refer to the supplementary.

\begin{table}[!t]
\centering
\small
    \begin{tabular}{c|ccc|c|c} 
    \hline
    \textbf{GTEA} & \multicolumn{3}{c|}{F1$@\{10,25,50\}$} & Edit & Acc \\ \hline
    LCDC~\cite{mac2019learning} & 75.4 & - & - & 72.8 & 65.3 \\ 
    TDRN~\cite{lei2018temporal} & 79.2 & 74.4 & 62.7 & 74.1 & 70.1 \\ 
    MS-TCN~\cite{farha2019ms}$\dag$ & 86.5 & 83.6 & 71.9 & 81.3 & 76.5 \\ 
    \hline
    SSTDA (65\%) & 85.2 & 82.6 & 69.3 & 79.6 & 75.7 \\ 
    \textbf{SSTDA} & \textbf{90.0} & \textbf{89.1} & \textbf{78.0} & \textbf{86.2} & \textbf{79.8} \\ \hline
    \hline
    \textbf{50Salads} & \multicolumn{3}{c|}{F1$@\{10,25,50\}$} & Edit & Acc \\ \hline
    TDRN~\cite{lei2018temporal} & 72.9 & 68.5 & 57.2 & 66.0 & 68.1 \\ 
    LCDC~\cite{mac2019learning} & 73.8 & - & - & 66.9 & 72.1 \\ 
    MS-TCN~\cite{farha2019ms}$\dag$ & 75.4 & 73.4 & 65.2 & 68.9 & 82.1 \\ \hline 
    SSTDA (65\%) & 77.7 & 75.0 & 66.2 & 69.3 & 80.7 \\ 
    \textbf{SSTDA} & \textbf{83.0} & \textbf{81.5} & \textbf{73.8} & \textbf{75.8} & \textbf{83.2} \\ \hline
    \hline
    \textbf{Breakfast} & \multicolumn{3}{c|}{F1$@\{10,25,50\}$} & Edit & Acc \\ \hline
    TCFPN~\cite{ding2018weakly} & - & - & - & - & 52.0 \\ 
    GRU~\cite{richard2017weakly} & - & - & - & - & 60.6 \\ 
    MS-TCN~\cite{farha2019ms}$\dag$ & 65.3 & 59.6 & 47.2 & 65.7 & 64.7 \\ 
    \hline
    SSTDA (65\%) & 69.3 & 62.9 & 49.4 & 69.0 & 65.8 \\ 
    \textbf{SSTDA} & \textbf{75.0} & \textbf{69.1} & \textbf{55.2} & \textbf{73.7} & \textbf{70.2} \\ \hline
    \end{tabular}
\caption{Comparison with the most recent action segmentation methods on all three datasets. SSTDA (65\%) means training with 65\% of total labeled training data. $\dag$Results from running the official code, as explained in \Cref{table:exp_local_global_DA}.} 

\label{table:SOTA}
\end{table}

\subsection{Ablation Study and Analysis}

\noindent\textbf{Design Choice for Local SSTDA:} \label{sec:design_local_SSTDA}
\begin{table}[!t]
\centering
\small
    \begin{tabular}{c|ccc|c|c} 
    \hline
     & \multicolumn{3}{c|}{F1$@\{10,25,50\}$} & Edit & Acc \\ \hline
    Source only & 86.5 & 83.6 & 71.9 & 81.3 & 76.5 \\ \hline
    $\{S1\}$ & 88.6 & 86.2 & 73.6 & 84.2 & 78.7 \\  
    $\{S2\}$ & 89.1 & 87.2 & \textbf{74.4} & 84.3 & 79.1 \\  
    $\{S3\}$ & 89.2 & 87.3 & 72.3 & 83.8 & 78.9 \\  
    $\{S4\}$ & 88.1 & 86.4 & 73.0 & 83.0 & 78.8 \\  \hline
    $\{S1, S2\}$ & 89.0 & 85.8 & 73.5 & \textbf{84.8} & 79.5 \\  
    $\{S2, S3\}$ & \textbf{89.6} & \textbf{87.9} & \textbf{74.4} & 84.5 & \textbf{80.1} \\  
    $\{S3, S4\}$ & 88.3 & 86.8 & 73.9 & 83.6 & 78.6 \\  \hline
    \end{tabular}
\caption{The experimental results of design choice for local SSTDA (on GTEA). $\{S_n\}$: add $\hat{G}_{ld}$ to the $n$th stage of MS-TCN, where smaller $n$ implies closer to inputs.} 
\label{table:stage_analysis}
\end{table}
Since we develop our approaches upon MS-TCN~\cite{farha2019ms}, it raises the question: \textit{How to effectively integrate binary domain prediction to a multi-stage architecture?}
To answer this, we first integrate $\hat{G}_{ld}$ into each stage and the results show that the best performance happens when the $\hat{G}_{ld}$ is integrated into middle stages, such as $S2$ or $S3$, as shown in \Cref{table:stage_analysis}. $S1$ is not a good choice for DA because it corresponds to low-level features with less discriminability where DA shows limited effects~\cite{long2015learning}, and represents less temporal receptive fields for videos.
However, higher stages (e.g. $S4$) are not always better. We conjecture that it is because the model fits more to the source data, causing difficulty for DA. In our case, integrating $\hat{G}_{ld}$ into $S2$ provides the best overall performance.

We also integrate binary domain prediction with multiple stages. However, multi-stage DA does not always guarantee improved performance. For example, $\{S1, S2\}$ has worse results than $\{S2\}$ in terms of F1$@\{10,25,50\}$. Since $\{S2\}$ and $\{S3\}$ provide the best single-stage DA performance, we use $\{S2, S3\}$, which performs the best, as the final model for all our approaches in all the experiments.

\noindent\textbf{Design Choice for Global SSTDA:} \label{sec:design_global_SSTDA}
\begin{table}[!t]
\centering
\small
    \begin{tabular}{c|ccc|c|c} 
    \hline
    Segment \# & \multicolumn{3}{c|}{F1$@\{10,25,50\}$} & Edit & Acc \\ \hline
    1 & 89.4 & 87.7 & 75.4 & 85.3 & 79.2 \\  
    \textbf{2} & \textbf{90.0} & \textbf{89.1} & \textbf{78.0} & \textbf{86.2} & \textbf{79.8} \\  
    3 & 89.7 & 87.6 & 75.4 & 85.2 & 79.2 \\  \hline
    \end{tabular}
\caption{The experimental results for different segment numbers of sequential domain prediction (on GTEA).} 
\label{table:segment_analysis}
\end{table}
The most critical design decision for the sequential domain prediction is the segment number for each video. In our implementation, we divide one source video into $m$ segments and do so for one target video, and then apply $G_{gd}$ to predict the permutation of domains for these $2m$ video segments. Therefore, the category number of $G_{gd}$ equals the number of all permutations $(2m)!/(m!)^2$. In other words, the segment number $m$ determine the complexity of the self-supervised auxiliary task. For example, $m=3$ leads to a 20-way classifier, and $m=4$ results in a 70-way classifier. Since a good self-supervised task should be neither naive nor over complicated~\cite{noroozi2016unsupervised}, we choose $m=2$ as our final decision, which is supported by our experiments as shown in \Cref{table:segment_analysis}.

\noindent\textbf{Segmentation Visualization:}
\begin{figure}[!t]
\centering
\includegraphics[width=0.475\textwidth]{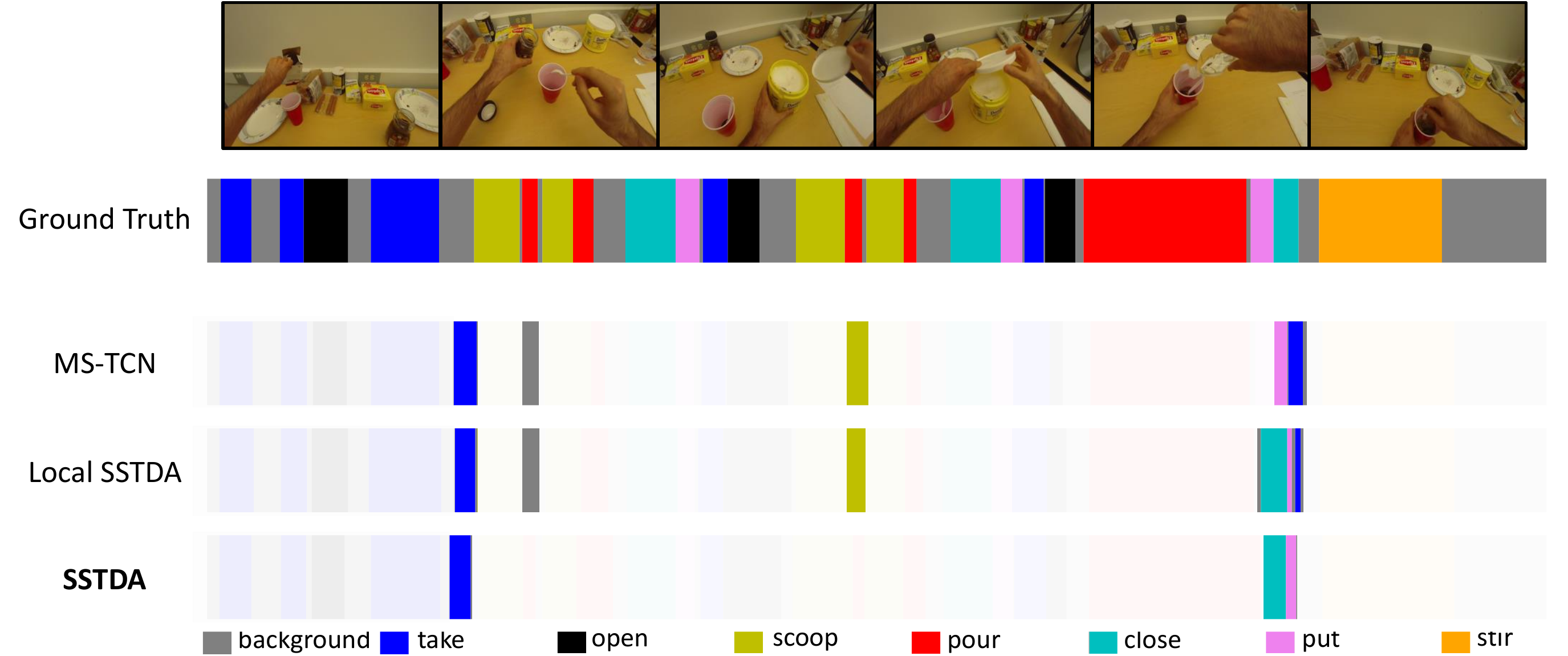}
\caption{The visualization of temporal action segmentation for our methods with color-coding (input example: \textit{make coffee}). 
``MS-TCN" is the baseline model without any DA methods.
We only highlight the action segments that are different from the ground truth for clear comparison.
}
\label{fig:qualitative_results_SSTDA}
\end{figure}
\begin{figure}[!t]
\centering
\includegraphics[width=0.475\textwidth]{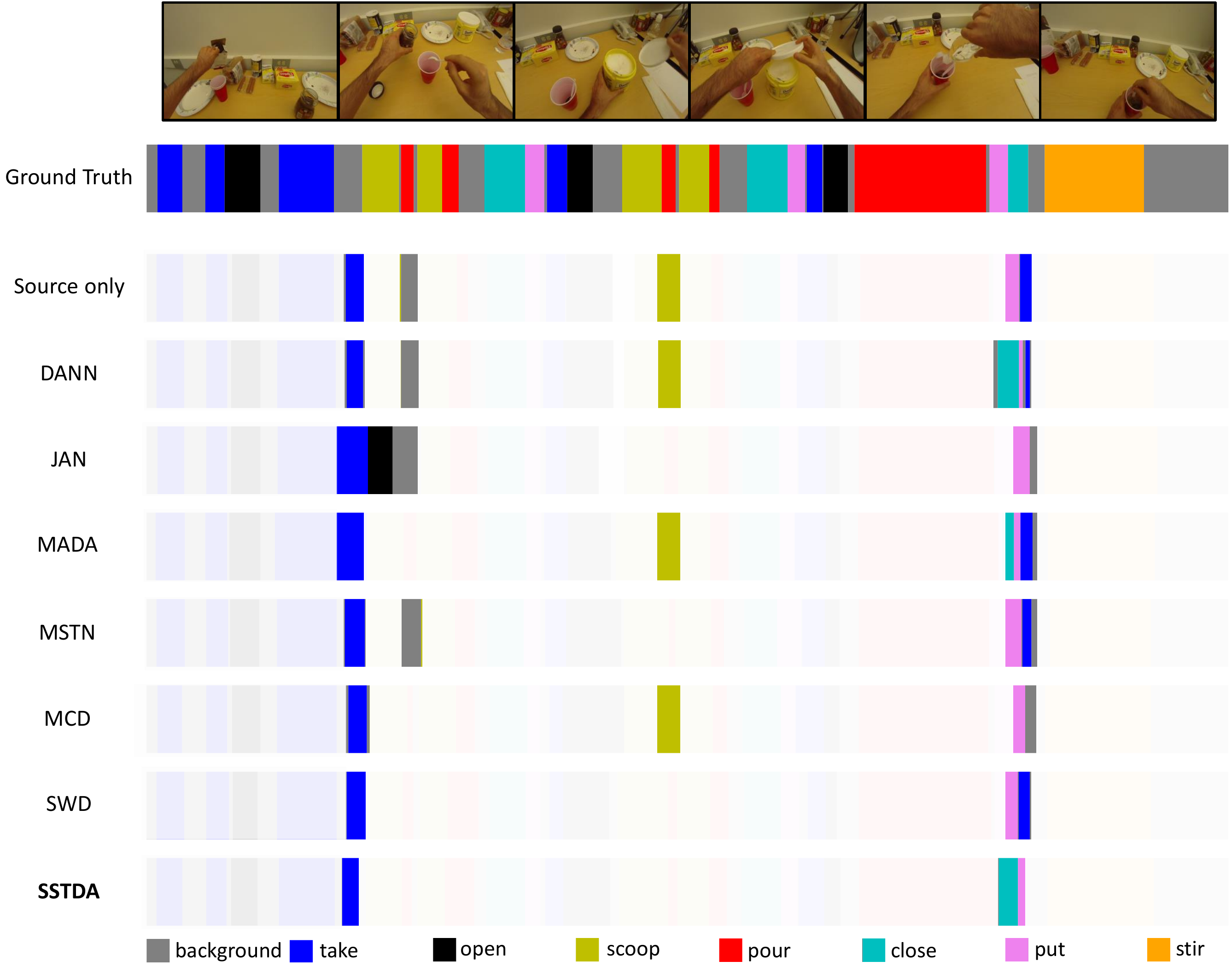}
\caption{The visualization of temporal action segmentation for different DA methods (same input as \Cref{fig:qualitative_results_SSTDA}).
``Source only" represents the baseline model, MS-TCN.
Only the segments different from the ground truth are highlighted.
}
\label{fig:qualitative_results_DA}
\end{figure}
It is also common to evaluate the qualitative performance to ensure that the prediction results are aligned with human vision. 
First, we compare our approaches with the baseline model MS-TCN~\cite{farha2019ms} and the ground truth, as shown in \Cref{fig:qualitative_results_SSTDA}.
MS-TCN fails to detect some \textit{pour} actions in the first half of the video, and falsely classify \textit{close} as \textit{take} in the latter part of the video. 
With local SSTDA, our approach can detect \textit{close} in the latter part of the video. 
Finally, with full SSTDA, our proposed method also detects all \textit{pour} action segments in the first half of video. 
We then compare SSTDA with other DA methods, and \Cref{fig:qualitative_results_DA} shows that our result is the closest to the ground truth. The others either incorrectly detect some actions or make incorrect classification. 
For more qualitative results, please refer to the supplementary.

\section{Conclusions and Future Work}
In this work, we propose a novel approach to effectively exploit unlabeled target videos to boost performance for action segmentation without target labels. To address the problem of spatio-temporal variations for videos across domains, 
we propose \textbf{Self-Supervised Temporal Domain Adaptation (SSTDA)} to jointly align cross-domain feature spaces embedded with local and global temporal dynamics by two self-supervised auxiliary tasks, \textit{binary} and \textit{sequential domain prediction}. 
Our experiments indicate that SSTDA outperforms other DA approaches by aligning temporal dynamics more effectively.
We also validate the proposed SSTDA on three challenging datasets (GTEA, 50Salads, and Breakfast), and show that SSTDA outperforms the current state-of-the-art method by large margins and only requires 65\% of the labeled training data to achieve the comparable performance, demonstrating the usefulness of adapting to unlabeled videos across variations. 
For the future work, we plan to apply SSTDA to more challenging video tasks (e.g. spatio-temporal action localization~\cite{gu2018ava}).

\section{Appendix}
In the supplementary material, we would like to show more details about the technical approach, implementation, and experiments.

\subsection{Technical Approach Details}

\noindent\textbf{Domain Attentive Temporal Pooling (DATP):}
\textit{Temporal pooling} is one of the most common methods to aggregate frame-level features into video-level features for each video. However, not all the frame-level features contribute the same to the overall domain discrepancy. Therefore, inspired by \cite{chen2019temporal,chen2020mixed}, we assign larger attention weights to the features which have larger domain discrepancy so that we can focus more on aligning those features, achieving more effective domain adaptation.

\begin{figure}[!t]
\centering
\includegraphics[width=0.475\textwidth]{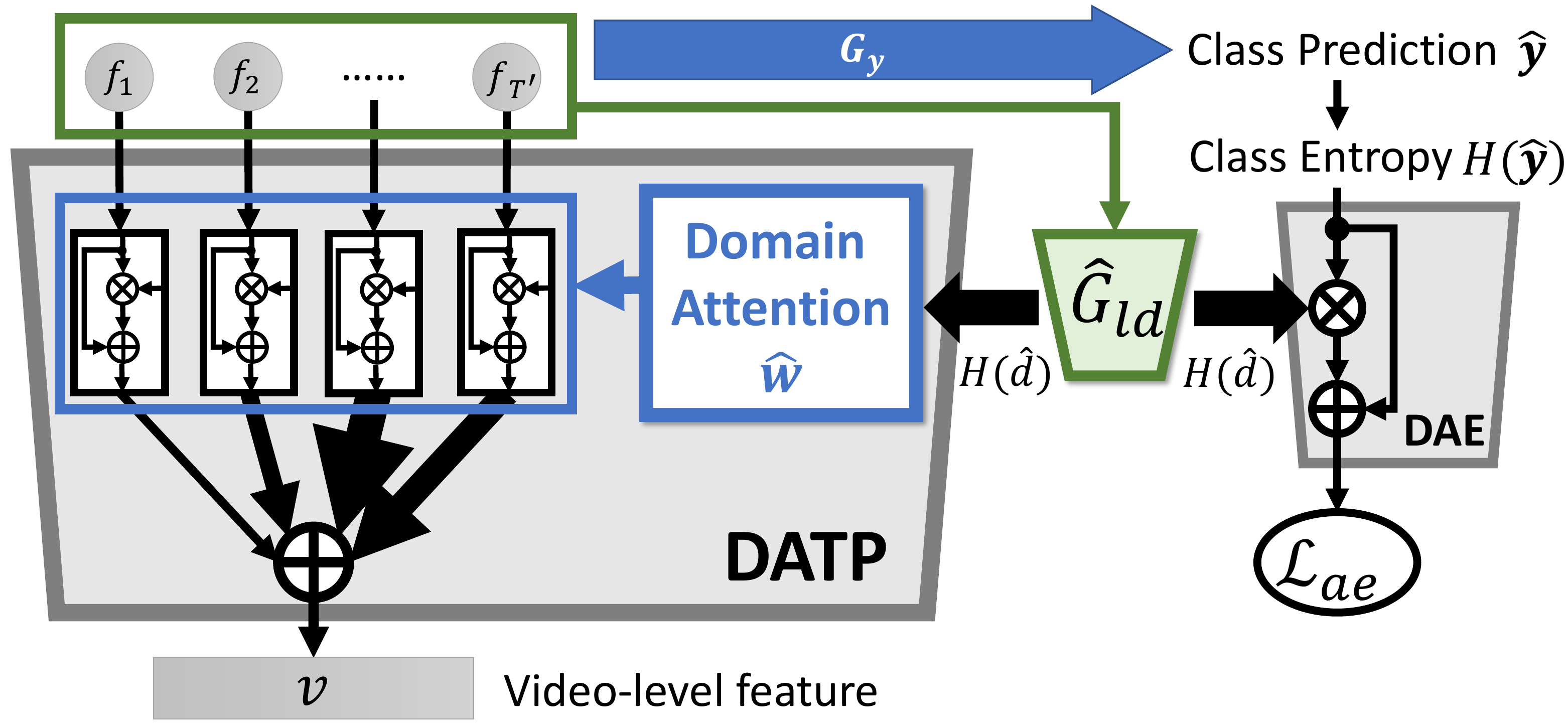}
\caption{The details of DATP (\textit{left}) and DAE (\textit{right}). Both modules take the domain entropy $H(\hat{d})$, which is calculated from the domain prediction $\hat{d}$, to calculate the attention weights. With the residual connection, DATP attends to the frame-level features for aggregating into the final video-level feature $v$ (arrow thickness represents assigned attention values), and DAE attends to the class entropy $H(\hat{\boldsymbol{y}})$ to obtain the attentive entropy loss $\mathcal{L}_{ae}$. }
\label{fig:DATP_DAE_supp}
\end{figure}

More specifically, we utilize the entropy criterion to generate the domain attention value for each frame-level feature $f_j$ as below:
\begin{equation} \label{eq:attention-weight_supp}
\small
\hat{w}_j = 1 - H(\hat{d}_j)
\end{equation}
where $\hat{d}_j$ is the output from the learned domain classifier $G_{ld}$ used in local SSTDA. $H(p) = -\sum_{k}p_k\cdot\log(p_k)$ is the entropy function to measure uncertainty. $\hat{w}_j$ increases when $H(\hat{d}_j)$ decreases, which means the domains can be distinguished well.
We also add a residual connection for more stable optimization. Finally, we aggregate the attended frame-level features with temporal pooling to generate the video-level feature $v$, which is noted as \textit{Domain Attentive Temporal Pooling (DATP)}, as illustrated in the left part of \Cref{fig:DATP_DAE_supp} and can be expressed as:
\begin{equation} \label{eq:attention-residual_supp}
\small
v = \frac{1}{T'}\sum_{j=1}^{T'}(\hat{w}_j + 1) \cdot f_j
\end{equation}
where $ + 1$ refers to the residual connection, and $\hat{w}_j + 1$ is equal to $w_j$ in the main paper. $T'$ is the number of frames used to generate a video-level feature. 

Local SSTDA is necessary to calculate the attention weights for DATP. Without this mechanisms, frames will be aggregated in the same way as \textit{temporal pooling} without cross-domain consideration, which is already demonstrated sub-optimal for cross-domain video tasks~\cite{chen2019temporal,chen2020mixed}.

\noindent\textbf{Domain Attentive Entropy (DAE):} 
Minimum entropy regularization is a common strategy to perform more refined classifier adaptation. However, we only want to minimize class entropy for the frames that are similar across domains. Therefore, inspired by \cite{wang2019transferable}, we attend to the frames which have low domain discrepancy, corresponding to high domain entropy $H(\hat{d}_j)$. More specifically, we adopt the \textit{Domain Attentive Entropy (DAE)} module to calculate the attentive entropy loss $\mathcal{L}_{ae}$, which can be expressed as follows:
\begin{equation} \label{eq:loss_attentive-entropy_supp}
\small
\mathcal{L}_{ae}=\frac{1}{T}\sum_{j=1}^{T}(H(\hat{d}_j)+1) \cdot H(\hat{y}_j)
\end{equation}
where $\hat{d}$ and $\hat{y}$ is the output of $\hat{G}_{ld}$ and $G_y$, respectively. $T$ is the total frame number of a video.
We also apply the residual connection for stability, as shown in the right part of \Cref{fig:DATP_DAE_supp}.

\noindent\textbf{Full Architecture:}
\begin{figure*}[!t]
\centering
\includegraphics[width=\textwidth]{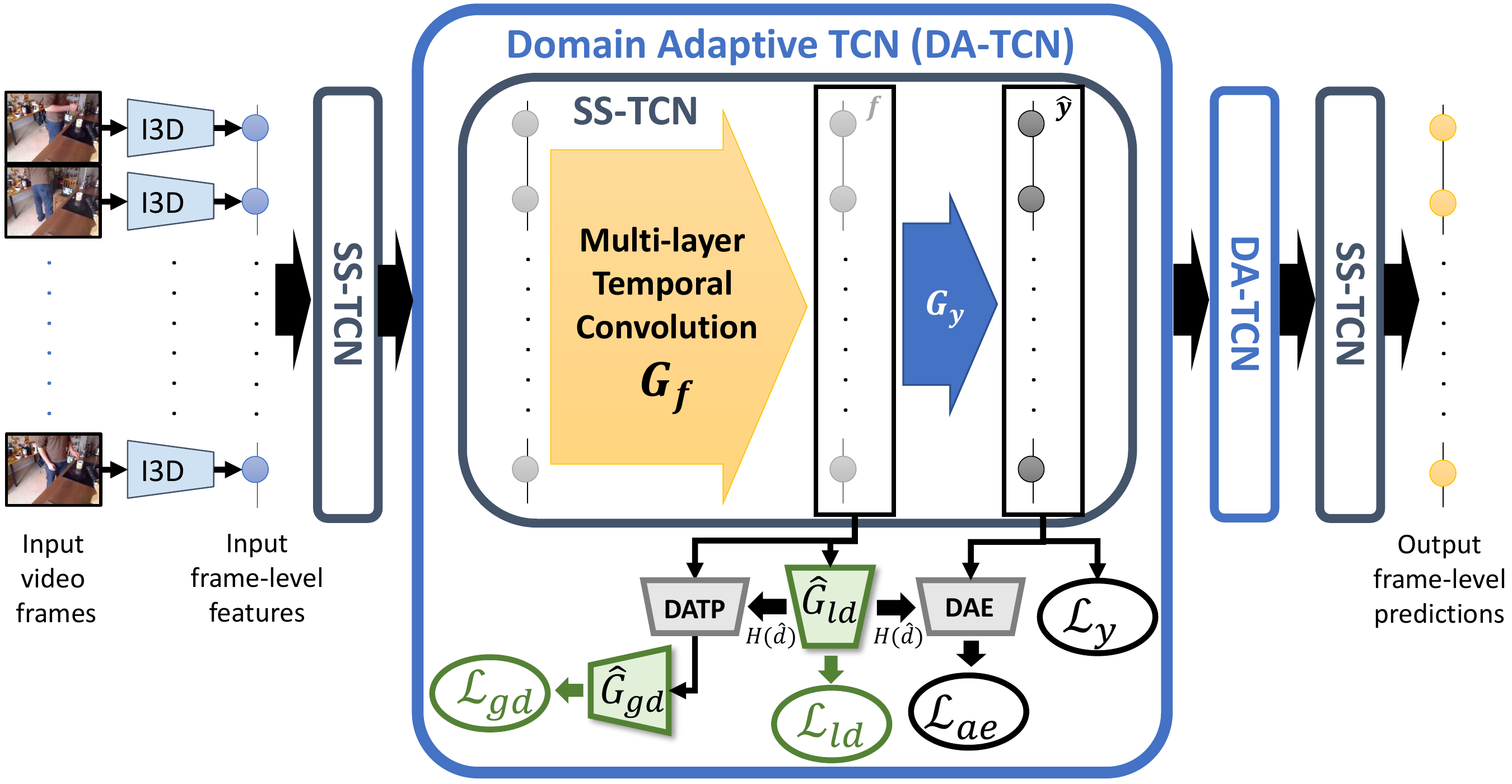}
\caption{The overall architecture of the proposed SSTDA. By equipping the network with a local adversarial domain classifier $\hat{G}_{ld}$, a global adversarial domain classifier $\hat{G}_{gd}$, a domain attentive temporal pooling (DATP) module, and a domain attentive entropy (DAE) module, we convert a SS-TCN into a DA-TCN, and stack multiple SS-TCNs and DA-TCNs to build the final architecture. $\mathcal{L}_{ld}$ and $\mathcal{L}_{gd}$ is the local and global domain loss, respectively. $\mathcal{L}_{y}$ is the prediction loss and $\mathcal{L}_{ae}$ is the attentive entropy loss. The domain entropy $H(\hat{d})$ is used to calculate the attention weights for DATP and DAE. An adversarial domain classifier $\hat{G}$ refers to a domain classifier $G$ equipped with a gradient reversal layer (GRL).
}
\label{fig:full_arch_supp}
\end{figure*}
Our method is built upon the state-of-the-art action segmentation model, MS-TCN~\cite{farha2019ms}, which takes input frame-level feature representations and generates the corresponding output frame-level class predictions by four stages of SS-TCN.
In our implementation, we convert the second and third stages into \textit{Domain Adaptive TCN (DA-TCN)} by integrating each SS-TCN with the following three parts: 1) $\hat{G}_{ld}$ (for \textit{binary domain prediction}), 2) DATP and $\hat{G}_{gd}$ (for \textit{sequential domain prediction}), and 3) DAE, bringing three corresponding loss functions, $\mathcal{L}_{ld}$, $\mathcal{L}_{gd}$ and $\mathcal{L}_{ae}$, respectively, as illustrated in \Cref{fig:full_arch_supp}.
The final loss function can be formulated as below:
\begin{equation} \label{eq:loss_SSTDA_supp}
\small
\begin{split}
\mathcal{L} = \sum^{4}_{s=1}\mathcal{L}_{y_{(s)}} - \sum^{3}_{s=2}(\beta_l\mathcal{L}_{ld_{(s)}}+\beta_g\mathcal{L}_{gd_{(s)}}-\mu\mathcal{L}_{ae_{(s)}})
\end{split}
\end{equation}
where $\beta_l$, $\beta_g$ and $\mu$ are the weights for $\mathcal{L}_{ld}$, $\mathcal{L}_{gd}$ and $\mathcal{L}_{ae}$, respectively, obtained by the methods described in \Cref{sec:optim_supp}. $s$ is the stage index in MS-TCN.

\subsection{Experiments}
\noindent\textbf{Datasets and Evaluation Metrics: }
\begin{table}[!t]
\centering
    \begin{tabular}{c|c|c|c}
     & \textbf{GTEA} & \textbf{50Salads} & \textbf{Breakfast} \\ \hline
    subject \# & 4 & 25 & 52 \\ \hline
    class \# & 11 & 17 & 48 \\ \hline
    video \# & 28 & 50 & 1712 \\ \hline
    avg. length (min.) & 1 & 6.4 & 2.7 \\ \hline
    avg. action \#/video & 20 & 20 & 6 \\ \hline
    cross-validation & 4-fold & 5-fold & 4-fold \\ \hline
    leave-\#-subject-out & 1 & 5 & 13 \\ \hline
    \end{tabular}
\caption{The statistics of action segmentation datasets.}
\label{table:dataset_supp}
\end{table}
The detailed statistics and the evaluation protocols of the three datasets are listed in \Cref{table:dataset_supp}. 
We follow \cite{lea2017temporal} to use the following three metrics for evaluation:
\begin{enumerate}
    \item \textit{Frame-wise accuracy (Acc)}: 
    Acc is one of the most typical evaluation metrics for action segmentation, but it does not consider the temporal dependencies of the prediction, causing the inconsistency between qualitative assessment and frame-wise accuracy. Besides, long action classes have higher impact on this metric than shorter action classes, making it not able to reflect over-segmentation errors. 
    \item \textit{Segmental edit score (Edit)}: 
    The edit score penalizes over-segmentation errors by measuring the ordering of predicted action segments independent of slight temporal shifts.
    \item \textit{Segmental F1 score at the IoU threshold $k\%$ (F1$@k$)}: 
    F1$@k$ also penalizes over-segmentation errors while ignoring minor temporal shifts between the predictions and ground truth.
    The scores are determined by the total number of actions but do not depend on the duration of each action instance, which is similar to mean average precision (mAP) with intersection-over-union (IoU) overlap criteria. F1$@k$ becomes popular recently since it better reflects the qualitative results.
\end{enumerate}

\noindent\textbf{Implementation and Optimization:} \label{sec:optim_supp}
Our implementation is based on the PyTorch~\cite{paszke2017automatic,steiner2019pytorch} framework. 
We extract I3D~\cite{carreira2017quo} features for the video frames and use these features as inputs to our model. The video frame rates are the same as \cite{farha2019ms}.
For GTEA and Breakfast datasets we use a video temporal resolution of 15 frames per second (fps), while for 50Salads we downsampled the features from 30 fps to 15 fps to be consistent with the other datasets.
For fair comparison, we adopt the same architecture design choices of MS-TCN~\cite{farha2019ms} as our baseline model. 
The whole model consists of four stages where each stage contains ten dilated convolution layers. We set the number of filters to $64$ in all
the layers of the model and the filter size is $3$. 
For optimization, we utilize the Adam optimizer and a batch size equal to $1$, following the official implementation of MS-TCN~\cite{farha2019ms}. Since the target data size is smaller than the source data, each target data is loaded randomly multiple times in each epoch during training.
For the weighting of loss functions, we follow the common strategy as \cite{ganin2015unsupervised,ganin2016domain} to gradually increase $\beta_l$ and $\beta_g$ from $0$ to $1$. The weighting $\alpha$ for smoothness loss is $0.15$ as in \cite{farha2019ms} and $\mu$ is chosen as $1 \times 10^{-2}$ via the grid-search.

\noindent\textbf{Less Training Labeled Data:}
\begin{table}[!t]
\centering
\small
    \begin{tabular}{c|c|ccc|c|c} 
    \hline
    \textbf{50Salads} & m\% & \multicolumn{3}{c|}{F1$@\{10,25,50\}$} & Edit & Acc \\ \hline
    \multirow{5}{*}{SSTDA} & 100\% & 83.0 & 81.5 &73.8 & 75.8 & 83.2 \\ 
    & 95\% & 81.6 & 80.0 & 73.1 & 75.6 & 83.2 \\ 
    & 85\% & 81.0 & 78.9 & 70.9 & 73.8 & 82.1 \\ 
    & 75\% & 78.9 & 76.5 & 68.6 & 71.7 & 81.1 \\ 
    & \textbf{65\%} & \textbf{77.7} & \textbf{75.0} & \textbf{66.2} & \textbf{69.3} & \textbf{80.7} \\ \hline
    MS-TCN & 100\% & 75.4 & 73.4 & 65.2 & 68.9 & 82.1 \\ 
    \hline
    \hline
    \textbf{GTEA} & m\% & \multicolumn{3}{c|}{F1$@\{10,25,50\}$} & Edit & Acc \\ \hline
    \multirow{2}{*}{SSTDA} & 100\% & 90.0 & 89.1 & 78.0 & 86.2 & 79.8 \\ 
     & \textbf{65\%} & \textbf{85.2} & \textbf{82.6} & \textbf{69.3} & \textbf{79.6} & \textbf{75.7} \\ \hline
    MS-TCN & 100\% & 86.5 & 83.6 & 71.9 & 81.3 & 76.5 \\ 
    \hline
    \hline
    \textbf{Breakfast} & m\% & \multicolumn{3}{c|}{F1$@\{10,25,50\}$} & Edit & Acc \\ \hline
    \multirow{2}{*}{SSTDA} & 100\% & 75.0 & 69.1 & 55.2 & 73.7 & 70.2 \\
     & \textbf{65\%} & \textbf{69.3} & \textbf{62.9} & \textbf{49.4} & \textbf{69.0} & \textbf{65.8} \\ \hline
    MS-TCN & 100\% & 65.3 & 59.6 & 47.2 & 65.7 & 64.7 \\ 
    \hline
    \end{tabular}
\caption{The comparison of SSTDA trained with less labeled training data. $m$ in the first row indicates the percentage of labeled training data used to train a model.} 
\label{table:less_training_data_supp}
\end{table}
To investigate the potential to train with a fewer number of labeled frames using SSTDA, we drop labeled frames from source domains with uniform sampling for training, and evaluate on the same length of validation data. Our experiment on the 50Salads dataset shows that by integrating with SSTDA, the performance does not drop significantly with the decrease in labeled training data, 
indicating the alleviation of reliance on labeled training data.
Finally, only \textit{65\%} of labeled training data are required to achieve comparable performance with MS-TCN, as shown in \Cref{table:less_training_data_supp}. 
We then evaluate the proposed SSTDA on GTEA and Breakfast with the same percentage of labeled training data, and also get comparable or better performance.

\Cref{table:less_training_data_supp} also indicates the results without additional labeled training data, which contain discriminative information that can directly boost the performance for action segmentation. 
The additional trained data are all unlabeled, so they cannot be directly trained with standard prediction loss. Therefore, we propose SSTDA to exploit unlabeled data to: 1) further improve the strong baseline, MS-TCN, without additional training labels, and 2) achieve comparable performance with this strong baseline using only 65\% of labels for training.

\noindent\textbf{Comparison with Other Approaches:}
\begin{table}[!t]
\centering
\small
    \begin{tabular}{c|ccc|c} 
    \hline
    \textbf{50Salads} & \multicolumn{3}{c|}{F1$@\{10,25,50\}$} & Edit \\ \hline
    Source only (MS-TCN) & 75.4 & 73.4 & 65.2 & 68.9 \\ \hline
    VCOP~\cite{xu2019self} & 75.8 & 73.8 & 65.9 & 68.4 \\ \hline
    DANN~\cite{ganin2016domain} & 79.2 & 77.8 & 70.3 & 72.0 \\ 
    JAN~\cite{long2017deep} & 80.9 & 79.4 & 72.4 & 73.5 \\ 
    MADA~\cite{pei2018multi} & 79.6 & 77.4 & 70.0 & 72.4 \\ 
    MSTN~\cite{xie2018learning} & 79.3 & 77.6 & 71.5 & 72.1 \\ 
    MCD~\cite{saito2018maximum} & 78.2 & 75.5 & 67.1 & 70.8 \\ 
    SWD~\cite{lee2019sliced} & 78.2 & 76.2 & 67.4 & 71.6 \\ \hline
    \textbf{SSTDA} & \textbf{83.0} & \textbf{81.5} & \textbf{73.8} & \textbf{75.8} \\ 
    \hline
    \hline
    \textbf{Breakfast} & \multicolumn{3}{c|}{F1$@\{10,25,50\}$} & Edit \\ \hline
    Source only (MS-TCN) & 65.3 & 59.6 & 47.2 & 65.7 \\ \hline
    VCOP~\cite{xu2019self} & 68.5 & 62.9 & 50.1 & 67.9 \\ \hline
    DANN~\cite{ganin2016domain} & 72.8 & 67.8 & 55.1 & 71.7 \\ 
    JAN~\cite{long2017deep} & 70.2 & 64.7 & 52.0 & 70.0 \\ 
    MADA~\cite{pei2018multi} & 71.0 & 65.4 & 52.8 & 71.2 \\ 
    MSTN~\cite{xie2018learning} & 69.6 & 63.6 & 51.5 & 69.2 \\ 
    MCD~\cite{saito2018maximum} & 70.4 & 65.1 & 52.4 & 69.7 \\ 
    SWD~\cite{lee2019sliced} & 68.6 & 63.2 & 50.6 & 69.1 \\ \hline
    \textbf{SSTDA} & \textbf{75.0} & \textbf{69.1} & \textbf{55.2} & \textbf{73.7} \\ 
    \hline
    \end{tabular}
\caption{The comparison of different methods that can learn information from unlabeled target videos (on 50Salads and Breakfast). All the methods are integrated with the same baseline model MS-TCN for fair comparison. 
} 
\label{table:exp_other_DA_supp}
\end{table}
We compare our proposed SSTDA with other approaches by integrating the same baseline architecture with other popular DA methods~\cite{ganin2016domain,long2017deep,pei2018multi,xie2018learning,saito2018maximum,lee2019sliced} and a state-of-the-art video-based self-supervised approach~\cite{xu2019self}. 
For fair comparison, all the methods are integrated with the second and third stages, as our proposed SSTDA, where the single-stage integration methods are described as follows:
\begin{enumerate}
    \item \textit{DANN~\cite{ganin2016domain}}: 
    We add one discriminator, which is the same as $G_{ld}$, equipped a gradient reversal layer (GRL) to the final frame-level features $\boldsymbol{f}$.
    
    \item \textit{JAN~\cite{long2017deep}}: 
    We integrate Joint Maximum Mean Discrepancy (JMMD) to the final frame-level features $\boldsymbol{f}$ and the class prediction $\boldsymbol{\hat{y}}$.
    
    \item \textit{MADA~\cite{pei2018multi}}: 
    Instead of a single discriminator, we add multiple discriminators according to the class number to calculate the domain loss for each class. All the class-based domain losses are weighted with prediction probabilities and then summed up to obtain the final domain loss.
    
    \item \textit{MSTN~\cite{xie2018learning}}: 
    We utilize pseudo-labels to cluster the data from the source and target domains, and calculate the class centroids for the source and target domain separately. Then we compute the semantic loss by calculating mean squared error (MSE) between the source and target centroids. The final loss contains the prediction loss, the semantic loss, and the domain loss as DANN~\cite{ganin2016domain}.
    
    \item \textit{MCD~\cite{saito2018maximum}}: 
    We apply another classifier $G'_y$ and follow the adversarial training procedure of Maximum Classifier Discrepancy to iteratively optimize the generator ($G_{f}$ in our case) and the classifier ($G_{y}$). The L1-distance is used as the discrepancy loss.
    
    \item \textit{SWD~\cite{lee2019sliced}}: 
    The framework is similar to MCD, but we replace the L1-distance with the Wasserstein distance as the discrepancy loss.
    
    \item \textit{VCOP~\cite{xu2019self}}: 
    We divide $\boldsymbol{f}$ into three segments and compute the segment-level features with temporal pooling. After temporal shuffling the segment-level features, pairwise features are computed and concatenated into the final feature representing the video clip order. The final features are then fed into a shallow classifier to predict the order.
    
\end{enumerate}

The experimental results on 50Salads and Breakfast both indicate that our proposed SSTDA outperforms all these methods, as shown in \Cref{table:exp_other_DA_supp}. 

The performance of the most recent video-based self-supervised learning method \cite{xu2019self} on 50Salads and Breakfast also show that temporal shuffling \textit{within single domain} without considering the relation across domains does not effectively benefit cross-domain action segmentation, resulting in even worse performance than other DA methods. 
Instead, our proposed self-supervised auxiliary tasks make predictions on cross-domain data, leading to cross-domain temporal relation reasoning instead
of predicting within-domain temporal orders, achieving significant improvement in the performance of our main task, action segmentation.

\subsection{Segmentation Visualization}

\begin{figure*}[!t]
\centering
    \begin{subfigure}[b]{\textwidth}
        \includegraphics[width=\textwidth]{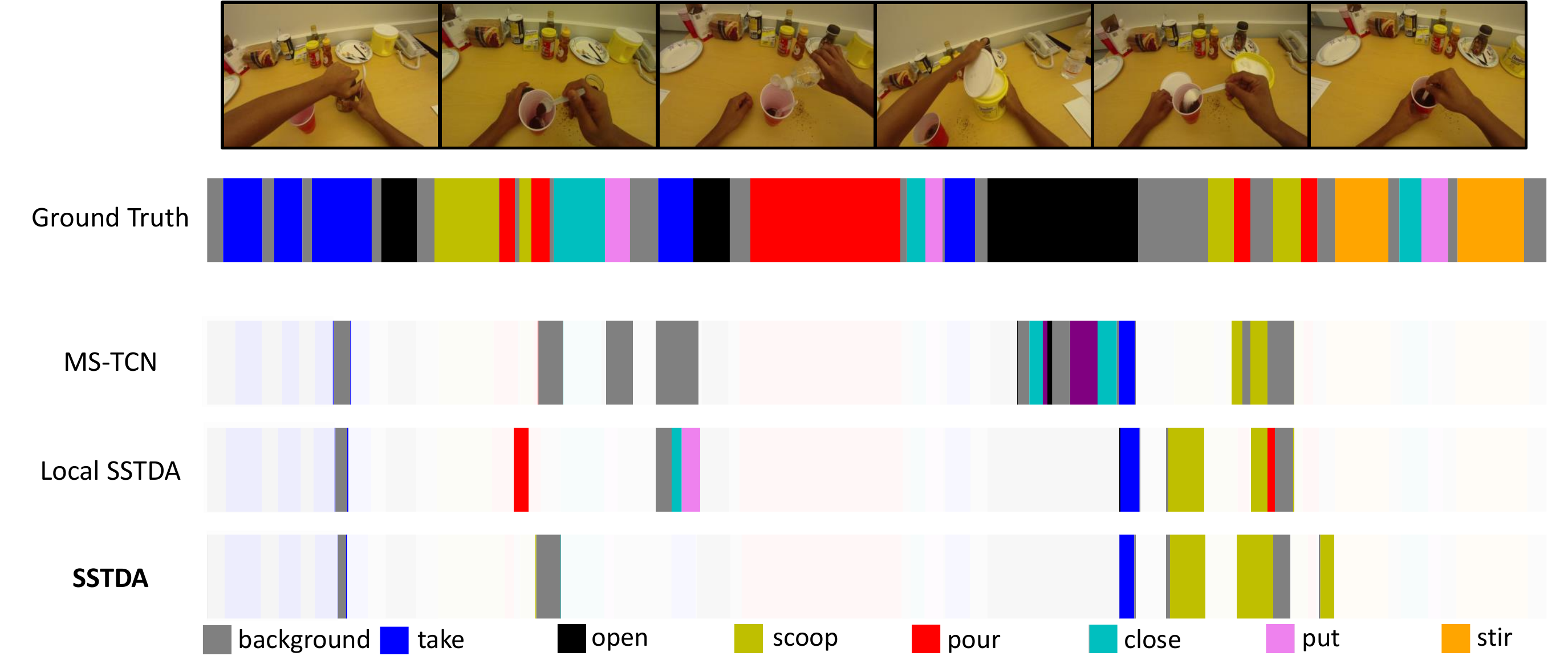}
        \caption{\textit{Make coffee}}
        \label{fig:qualitative_S2_Coffee_supp}
    \end{subfigure}
    \begin{subfigure}[b]{\textwidth}
        \includegraphics[width=\textwidth]{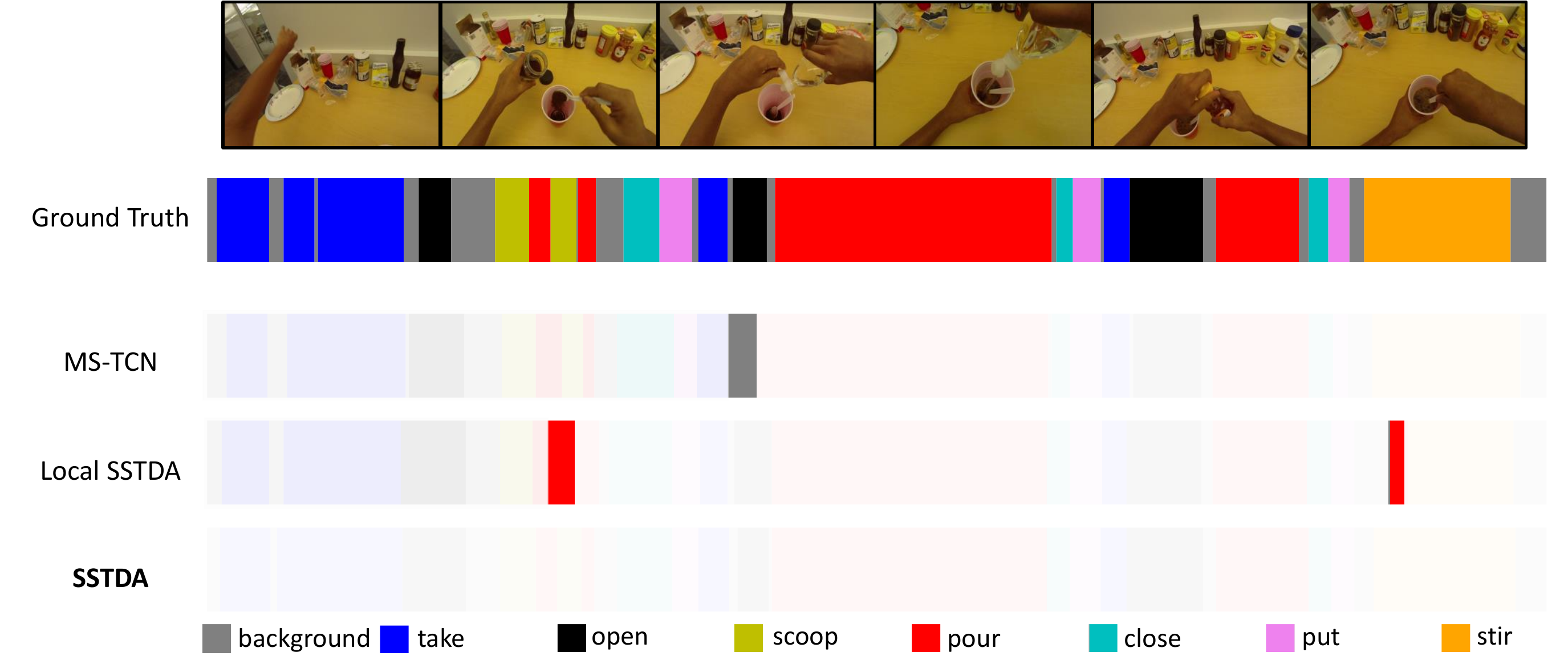}
        \caption{\textit{Make honey coffee}}
        \label{fig:qualitative_S2_CofHoney_supp}
    \end{subfigure}
\caption{The visualization of temporal action segmentation for our methods with color-coding on \textit{GTEA}. 
The video snapshots and the segmentation visualization are in the same temporal order (from left to right). 
We only highlight the action segments that are significantly different from the ground truth for clear comparison. 
``MS-TCN" represents the baseline model trained with only source data.
}
\label{fig:qualitative_GTEA_supp}
\end{figure*}

\begin{figure*}[!t]
\centering
    \begin{subfigure}[b]{\textwidth}
        \includegraphics[width=\textwidth]{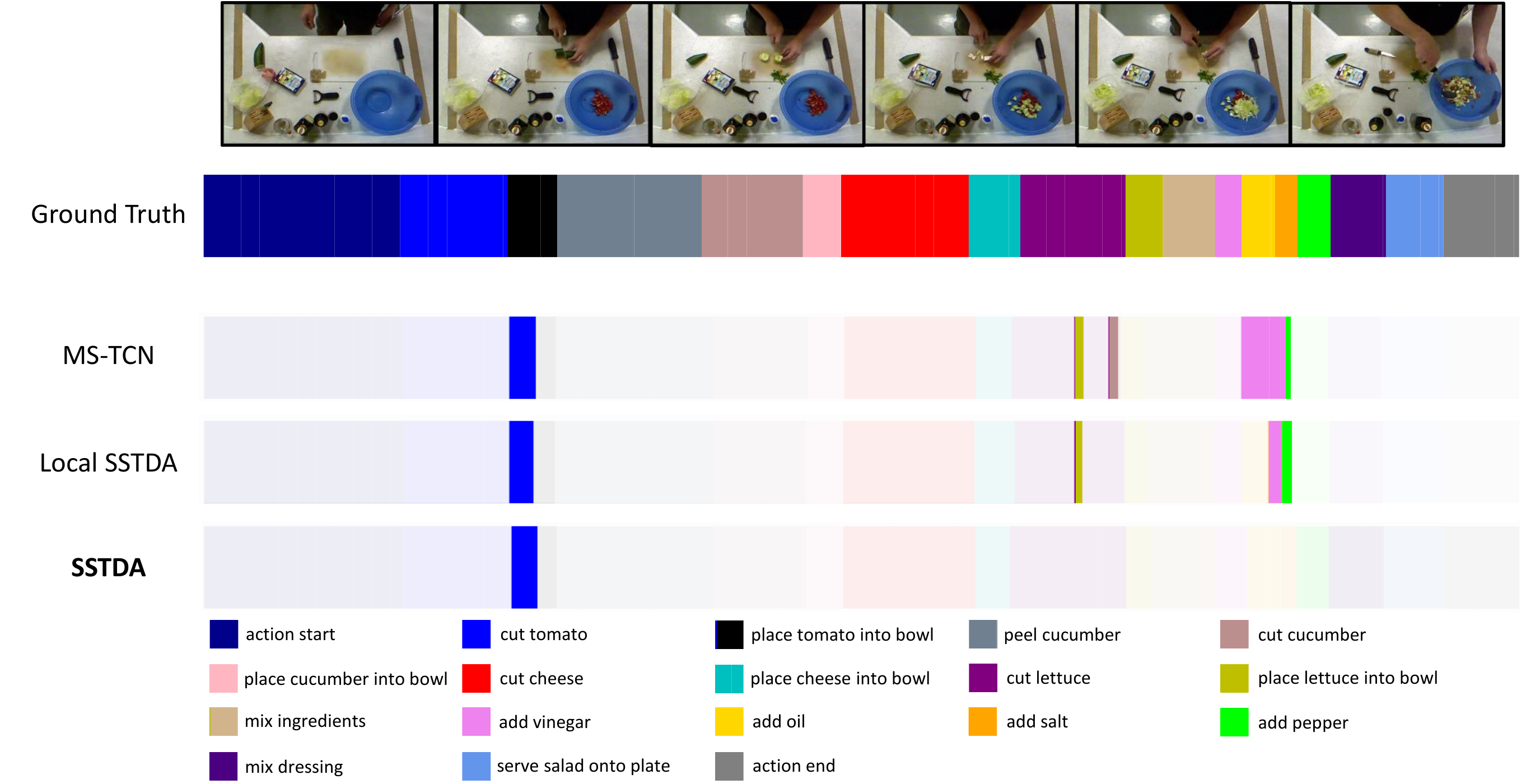}
        \caption{\textit{Subject 02}}
        \label{fig:qualitative_rgb-02-2_supp}
    \end{subfigure}
    \begin{subfigure}[b]{\textwidth}
        \includegraphics[width=\textwidth]{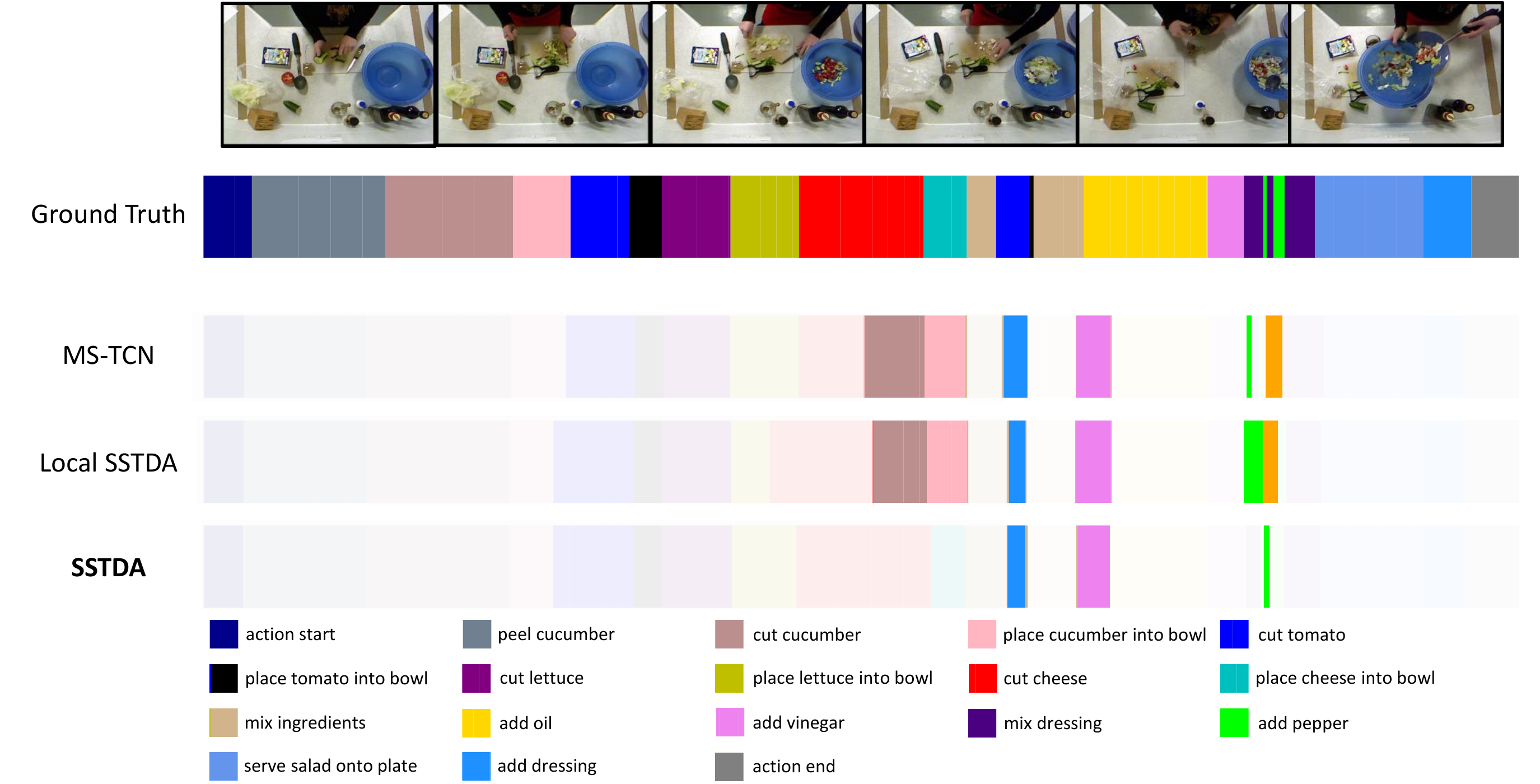}
        \caption{\textit{Subject 04}}
        \label{fig:qualitative_rgb-04-2_supp}
    \end{subfigure}
\caption{The visualization of temporal action segmentation for our methods with color-coding on \textit{50Salads}. 
The video snapshots and the segmentation visualization are in the same temporal order (from left to right). 
We only highlight the action segments that are different from the ground truth for clear comparison. 
Both examples correspond to the same activity \textit{Make salad}, but they are performed by different subjects, i.e., people.
}
\label{fig:qualitative_50Salads_supp}
\end{figure*}

\begin{figure*}[!t]
\centering
    \begin{subfigure}[b]{\textwidth}
        \includegraphics[width=\textwidth]{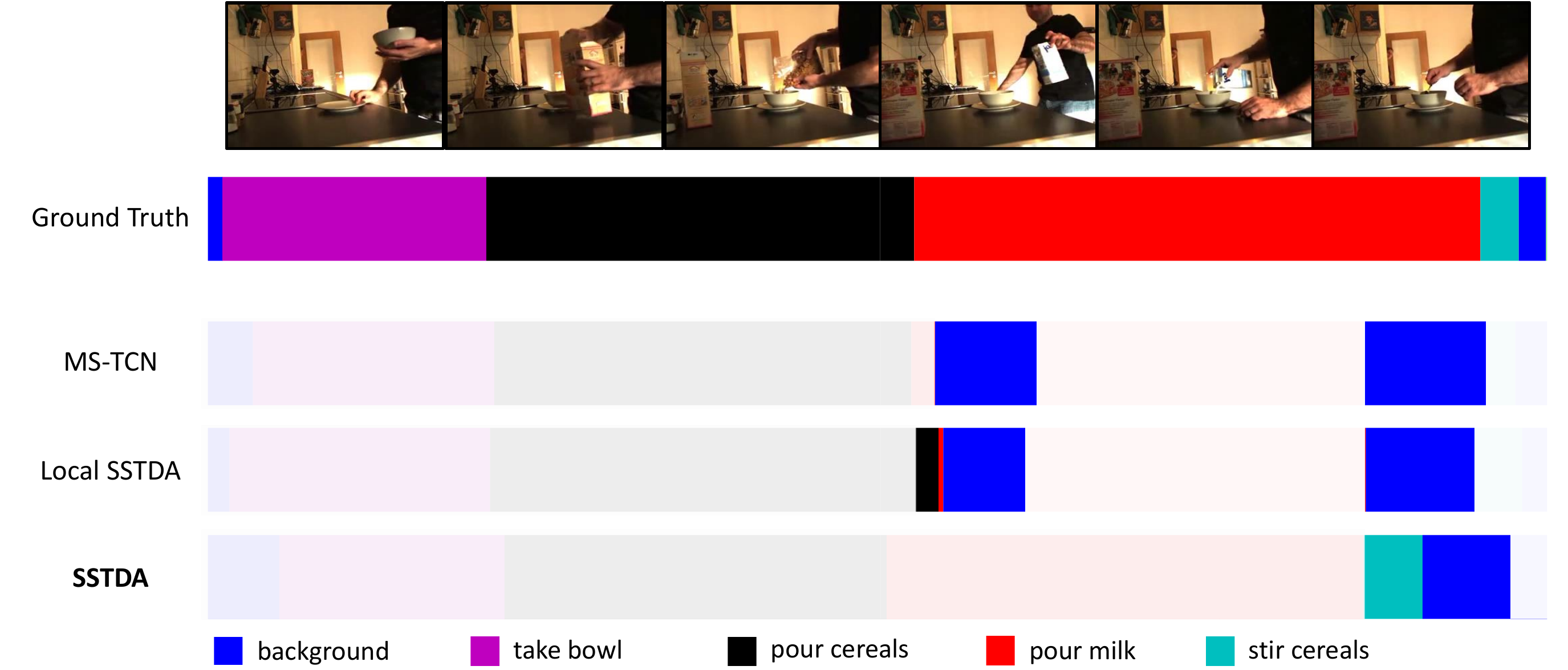} 
        \caption{\textit{Make Cereal}}
        \label{fig:qualitative_P05_cam01_cereals_supp}
    \end{subfigure}
    \begin{subfigure}[b]{\textwidth}
        \includegraphics[width=\textwidth]{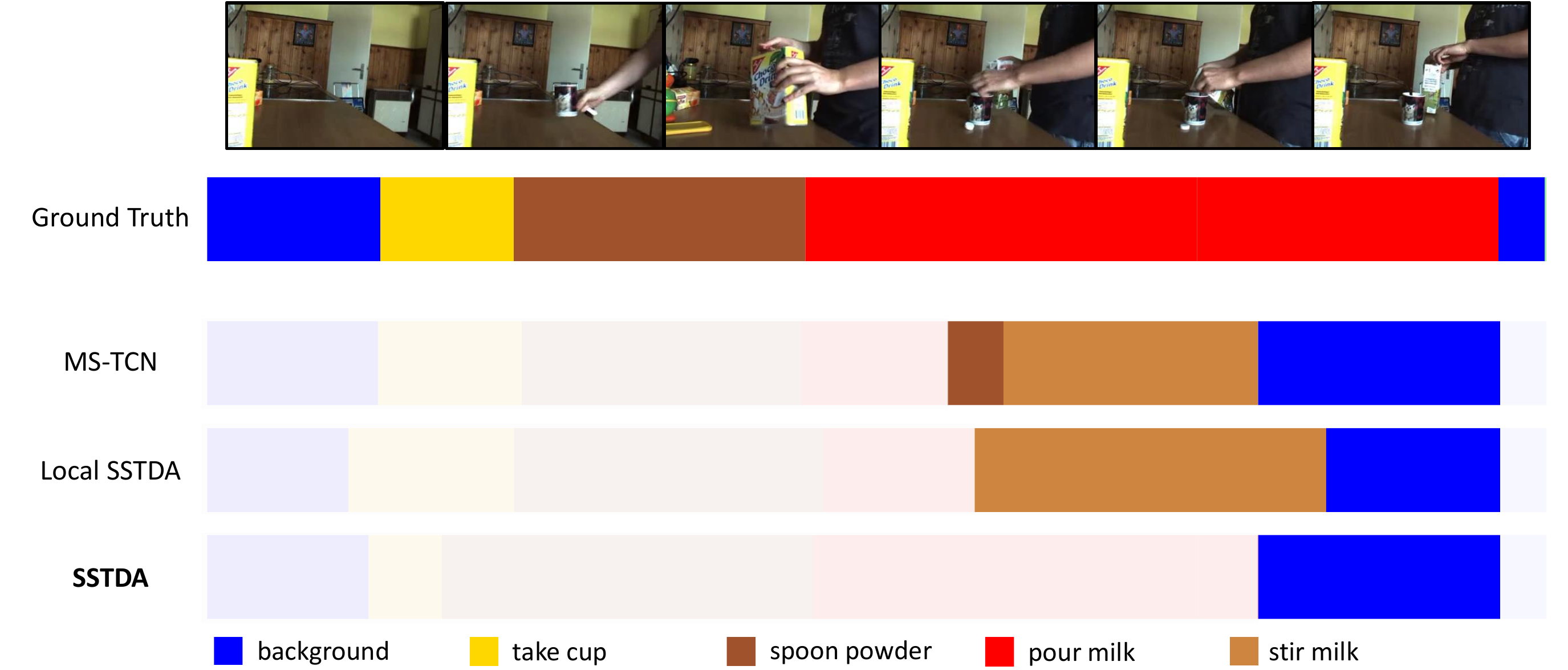}
        \caption{\textit{Make milk}}
        \label{fig:qualitative_P07_cam01_milk_supp}
    \end{subfigure}
\caption{The visualization of temporal action segmentation for our methods with color-coding on \textit{Breakfast}. 
The video snapshots and the segmentation visualization are in the same temporal order (from left to right). 
We only highlight the action segments that are different from the ground truth for clear comparison. 
}
\label{fig:qualitative_Breakfast_supp}
\end{figure*}

Here we show more qualitative segmentation results from all three datasets to compare our methods with the baseline model, MS-TCN~\cite{farha2019ms}.
All the results (\Cref{fig:qualitative_GTEA_supp} for \textit{GTEA}, \Cref{fig:qualitative_50Salads_supp} for \textit{50Salads}, and \Cref{fig:qualitative_Breakfast_supp} for \textit{Breakfast}) demonstrate that the improvement over the baseline by only local SSTDA is sometimes limited. 
For example, local SSTDA falsely detects the \textit{pour} action in \Cref{fig:qualitative_S2_CofHoney_supp}, falsely classifies \textit{cheese}-related actions as \textit{cucumber}-related actions in \Cref{fig:qualitative_rgb-04-2_supp}, and falsely detects the \textit{stir milk} action in \Cref{fig:qualitative_P07_cam01_milk_supp}.
However, by jointly aligning local and global temporal dynamics with SSTDA, the model is effectively adapted to the target domain, reducing the above mentioned incorrect predictions and achieving better segmentation.


{\small
\bibliographystyle{ieee_fullname}
\bibliography{egbib}

\begin{thebibliography}{10}\itemsep=-1pt

\bibitem{ahsan2019video}
Unaiza Ahsan, Rishi Madhok, and Irfan Essa.
\newblock Video jigsaw: Unsupervised learning of spatiotemporal context for
  video action recognition.
\newblock In {\em IEEE Winter Conference on Applications of Computer Vision
  (WACV)}, 2019.

\bibitem{carreira2017quo}
Joao Carreira and Andrew Zisserman.
\newblock Quo vadis, action recognition? a new model and the kinetics dataset.
\newblock In {\em IEEE conference on Computer Vision and Pattern Recognition
  (CVPR)}, 2017.

\bibitem{chen2019taaan}
Min-Hung Chen, Zsolt Kira, and Ghassan AlRegib.
\newblock Temporal attentive alignment for video domain adaptation.
\newblock {\em CVPR Workshop on Learning from Unlabeled Videos}, 2019.

\bibitem{chen2019temporal}
Min-Hung Chen, Zsolt Kira, Ghassan AlRegib, Jaekwon Woo, Ruxin Chen, and Jian
  Zheng.
\newblock Temporal attentive alignment for large-scale video domain adaptation.
\newblock In {\em IEEE International Conference on Computer Vision (ICCV)},
  2019.

\bibitem{chen2020mixed}
Min-Hung Chen, Baopu Li, Yingze Bao, and Ghassan AlRegib.
\newblock Action segmentation with mixed temporal domain adaptation.
\newblock In {\em The IEEE Winter Conference on Applications of Computer Vision
  (WACV)}, 2020.

\bibitem{csurka2017comprehensive}
Gabriela Csurka.
\newblock A comprehensive survey on domain adaptation for visual applications.
\newblock In {\em Domain Adaptation in Computer Vision Applications}, pages
  1--35. Springer, 2017.

\bibitem{ding2017tricornet}
Li Ding and Chenliang Xu.
\newblock Tricornet: A hybrid temporal convolutional and recurrent network for
  video action segmentation.
\newblock {\em arXiv preprint arXiv:1705.07818}, 2017.

\bibitem{ding2018weakly}
Li Ding and Chenliang Xu.
\newblock Weakly-supervised action segmentation with iterative soft boundary
  assignment.
\newblock In {\em IEEE Conference on Computer Vision and Pattern Recognition
  (CVPR)}, 2018.

\bibitem{farha2019ms}
Yazan~Abu Farha and Jurgen Gall.
\newblock Ms-tcn: Multi-stage temporal convolutional network for action
  segmentation.
\newblock In {\em IEEE Conference on Computer Vision and Pattern Recognition
  (CVPR)}, 2019.

\bibitem{fathi2011learning}
Alireza Fathi, Xiaofeng Ren, and James~M Rehg.
\newblock Learning to recognize objects in egocentric activities.
\newblock In {\em IEEE Conference on Computer Vision and Pattern Recognition
  (CVPR)}, 2011.

\bibitem{ganin2015unsupervised}
Yaroslav Ganin and Victor Lempitsky.
\newblock Unsupervised domain adaptation by backpropagation.
\newblock In {\em International Conference on Machine Learning (ICML)}, 2015.

\bibitem{ganin2016domain}
Yaroslav Ganin, Evgeniya Ustinova, Hana Ajakan, Pascal Germain, Hugo
  Larochelle, Fran{\c{c}}ois Laviolette, Mario Marchand, and Victor Lempitsky.
\newblock Domain-adversarial training of neural networks.
\newblock {\em The Journal of Machine Learning Research (JMLR)},
  17(1):2096--2030, 2016.

\bibitem{goodfellow2014generative}
Ian Goodfellow, Jean Pouget-Abadie, Mehdi Mirza, Bing Xu, David Warde-Farley,
  Sherjil Ozair, Aaron Courville, and Yoshua Bengio.
\newblock Generative adversarial nets.
\newblock In {\em Advances in Neural Information Processing Systems (NeurIPS)},
  2014.

\bibitem{gu2018ava}
Chunhui Gu, Chen Sun, David~A Ross, Carl Vondrick, Caroline Pantofaru, Yeqing
  Li, Sudheendra Vijayanarasimhan, George Toderici, Susanna Ricco, Rahul
  Sukthankar, et~al.
\newblock Ava: A video dataset of spatio-temporally localized atomic visual
  actions.
\newblock In {\em IEEE Conference on Computer Vision and Pattern Recognition
  (CVPR)}, 2018.

\bibitem{jamal2018deep}
Arshad Jamal, Vinay~P Namboodiri, Dipti Deodhare, and KS Venkatesh.
\newblock Deep domain adaptation in action space.
\newblock In {\em British Machine Vision Conference (BMVC)}, 2018.

\bibitem{kim2019self}
Dahun Kim, Donghyeon Cho, and In~So Kweon.
\newblock Self-supervised video representation learning with space-time cubic
  puzzles.
\newblock In {\em AAAI Conference on Artificial Intelligence (AAAI)}, 2019.

\bibitem{kong2018human}
Yu Kong and Yun Fu.
\newblock Human action recognition and prediction: A survey.
\newblock {\em arXiv preprint arXiv:1806.11230}, 2018.

\bibitem{kuehne2014language}
Hilde Kuehne, Ali Arslan, and Thomas Serre.
\newblock The language of actions: Recovering the syntax and semantics of
  goal-directed human activities.
\newblock In {\em IEEE Conference on Computer Vision and Pattern Recognition
  (CVPR)}, 2014.

\bibitem{kurmi2019attending}
Vinod~Kumar Kurmi, Shanu Kumar, and Vinay~P Namboodiri.
\newblock Attending to discriminative certainty for domain adaptation.
\newblock In {\em IEEE conference on Computer Vision and Pattern Recognition
  (CVPR)}, 2019.

\bibitem{lahiri2019unsupervised}
Avisek Lahiri, Sri~Charan Ragireddy, Prabir Biswas, and Pabitra Mitra.
\newblock Unsupervised adversarial visual level domain adaptation for learning
  video object detectors from images.
\newblock In {\em IEEE Winter Conference on Applications of Computer Vision
  (WACV)}, 2019.

\bibitem{lea2017temporal}
Colin Lea, Michael~D Flynn, Rene Vidal, Austin Reiter, and Gregory~D Hager.
\newblock Temporal convolutional networks for action segmentation and
  detection.
\newblock In {\em IEEE Conference on Computer Vision and Pattern Recognition
  (CVPR)}, 2017.

\bibitem{lee2019sliced}
Chen-Yu Lee, Tanmay Batra, Mohammad~Haris Baig, and Daniel Ulbricht.
\newblock Sliced wasserstein discrepancy for unsupervised domain adaptation.
\newblock In {\em IEEE conference on Computer Vision and Pattern Recognition
  (CVPR)}, 2019.

\bibitem{lee2017unsupervised}
Hsin-Ying Lee, Jia-Bin Huang, Maneesh Singh, and Ming-Hsuan Yang.
\newblock Unsupervised representation learning by sorting sequences.
\newblock In {\em IEEE International Conference on Computer Vision (ICCV)},
  2017.

\bibitem{lei2018temporal}
Peng Lei and Sinisa Todorovic.
\newblock Temporal deformable residual networks for action segmentation in
  videos.
\newblock In {\em IEEE Conference on Computer Vision and Pattern Recognition
  (CVPR)}, 2018.

\bibitem{long2015learning}
Mingsheng Long, Yue Cao, Jianmin Wang, and Michael Jordan.
\newblock Learning transferable features with deep adaptation networks.
\newblock In {\em International Conference on Machine Learning (ICML)}, 2015.

\bibitem{long2016unsupervised}
Mingsheng Long, Han Zhu, Jianmin Wang, and Michael~I Jordan.
\newblock Unsupervised domain adaptation with residual transfer networks.
\newblock In {\em Advances in Neural Information Processing Systems (NeurIPS)},
  2016.

\bibitem{long2017deep}
Mingsheng Long, Han Zhu, Jianmin Wang, and Michael~I Jordan.
\newblock Deep transfer learning with joint adaptation networks.
\newblock In {\em International Conference on Machine Learning (ICML)}, 2017.

\bibitem{ma2019ts}
Chih-Yao Ma, Min-Hung Chen, Zsolt Kira, and Ghassan AlRegib.
\newblock Ts-lstm and temporal-inception: Exploiting spatiotemporal dynamics
  for activity recognition.
\newblock {\em Signal Processing: Image Communication}, 71:76--87, 2019.

\bibitem{ma2018attend}
Chih-Yao Ma, Asim Kadav, Iain Melvin, Zsolt Kira, Ghassan AlRegib, and
  Hans~Peter Graf.
\newblock Attend and interact: higher-order object interactions for video
  understanding.
\newblock In {\em IEEE conference on Computer Vision and Pattern Recognition
  (CVPR)}, 2018.

\bibitem{mac2019learning}
Khoi-Nguyen~C Mac, Dhiraj Joshi, Raymond~A Yeh, Jinjun Xiong, Rogerio~S Feris,
  and Minh~N Do.
\newblock Learning motion in feature space: Locally-consistent deformable
  convolution networks for fine-grained action detection.
\newblock In {\em IEEE International Conference on Computer Vision (ICCV)},
  2019.

\bibitem{noroozi2016unsupervised}
Mehdi Noroozi and Paolo Favaro.
\newblock Unsupervised learning of visual representations by solving jigsaw
  puzzles.
\newblock In {\em European Conference on Computer Vision (ECCV)}, 2016.

\bibitem{pan2010survey}
Sinno~Jialin Pan, Qiang Yang, et~al.
\newblock A survey on transfer learning.
\newblock {\em IEEE Transactions on Knowledge and Data Engineering (TKDE)},
  22(10):1345--1359, 2010.

\bibitem{paszke2017automatic}
Adam Paszke, Sam Gross, Soumith Chintala, Gregory Chanan, Edward Yang, Zachary
  DeVito, Zeming Lin, Alban Desmaison, Luca Antiga, and Adam Lerer.
\newblock Automatic differentiation in pytorch.
\newblock In {\em Advances in Neural Information Processing Systems Workshop
  (NeurIPSW)}, 2017.

\bibitem{pei2018multi}
Zhongyi Pei, Zhangjie Cao, Mingsheng Long, and Jianmin Wang.
\newblock Multi-adversarial domain adaptation.
\newblock In {\em AAAI Conference on Artificial Intelligence (AAAI)}, 2018.

\bibitem{richard2017weakly}
Alexander Richard, Hilde Kuehne, and Juergen Gall.
\newblock Weakly supervised action learning with rnn based fine-to-coarse
  modeling.
\newblock In {\em IEEE Conference on Computer Vision and Pattern Recognition
  (CVPR)}, 2017.

\bibitem{saito2018maximum}
Kuniaki Saito, Kohei Watanabe, Yoshitaka Ushiku, and Tatsuya Harada.
\newblock Maximum classifier discrepancy for unsupervised domain adaptation.
\newblock In {\em IEEE conference on Computer Vision and Pattern Recognition
  (CVPR)}, 2018.

\bibitem{stein2013combining}
Sebastian Stein and Stephen~J McKenna.
\newblock Combining embedded accelerometers with computer vision for
  recognizing food preparation activities.
\newblock In {\em ACM international joint conference on Pervasive and
  ubiquitous computing (UbiComp)}, 2013.

\bibitem{steiner2019pytorch}
Benoit Steiner, Zachary DeVito, Soumith Chintala, Sam Gross, Adam Paszke,
  Francisco Massa, Adam Lerer, Gregory Chanan, Zeming Lin, Edward Yang, et~al.
\newblock Pytorch: An imperative style, high-performance deep learning library.
\newblock {\em Advances in Neural Information Processing Systems (NeurIPS)},
  2019.

\bibitem{sultani2014human}
Waqas Sultani and Imran Saleemi.
\newblock Human action recognition across datasets by foreground-weighted
  histogram decomposition.
\newblock In {\em IEEE conference on Computer Vision and Pattern Recognition
  (CVPR)}, 2014.

\bibitem{tzeng2017adversarial}
Eric Tzeng, Judy Hoffman, Kate Saenko, and Trevor Darrell.
\newblock Adversarial discriminative domain adaptation.
\newblock In {\em IEEE Conference on Computer Vision and Pattern Recognition
  (CVPR)}, 2017.

\bibitem{wang2018non}
Xiaolong Wang, Ross Girshick, Abhinav Gupta, and Kaiming He.
\newblock Non-local neural networks.
\newblock In {\em IEEE conference on Computer Vision and Pattern Recognition
  (CVPR)}, 2018.

\bibitem{wang2019transferable}
Ximei Wang, Liang Li, Weirui Ye, Mingsheng Long, and Jianmin Wang.
\newblock Transferable attention for domain adaptation.
\newblock In {\em AAAI Conference on Artificial Intelligence (AAAI)}, 2019.

\bibitem{wei2018learning}
Donglai Wei, Joseph~J Lim, Andrew Zisserman, and William~T Freeman.
\newblock Learning and using the arrow of time.
\newblock In {\em IEEE Conference on Computer Vision and Pattern Recognition
  (CVPR)}, 2018.

\bibitem{xie2018learning}
Shaoan Xie, Zibin Zheng, Liang Chen, and Chuan Chen.
\newblock Learning semantic representations for unsupervised domain adaptation.
\newblock In {\em International Conference on Machine Learning (ICML)}, 2018.

\bibitem{xu2019self}
Dejing Xu, Jun Xiao, Zhou Zhao, Jian Shao, Di Xie, and Yueting Zhuang.
\newblock Self-supervised spatiotemporal learning via video clip order
  prediction.
\newblock In {\em IEEE Conference on Computer Vision and Pattern Recognition
  (CVPR)}, 2019.

\bibitem{xu2016dual}
Tiantian Xu, Fan Zhu, Edward~K Wong, and Yi Fang.
\newblock Dual many-to-one-encoder-based transfer learning for cross-dataset
  human action recognition.
\newblock {\em Image and Vision Computing}, 55:127--137, 2016.

\bibitem{zhang2018collaborative}
Weichen Zhang, Wanli Ouyang, Wen Li, and Dong Xu.
\newblock Collaborative and adversarial network for unsupervised domain
  adaptation.
\newblock In {\em IEEE Conference on Computer Vision and Pattern Recognition
  (CVPR)}, 2018.

\bibitem{zhang2019learning}
Xiao-Yu Zhang, Haichao Shi, Changsheng Li, Kai Zheng, Xiaobin Zhu, and Lixin
  Duan.
\newblock Learning transferable self-attentive representations for action
  recognition in untrimmed videos with weak supervision.
\newblock In {\em AAAI Conference on Artificial Intelligence (AAAI)}, 2019.

\end{thebibliography}
}

\end{document}